%% file: acl_latex.tex
\pdfoutput=1

\documentclass[11pt]{article}

\usepackage[preprint]{acl}
\usepackage{graphicx}
\usepackage{booktabs}
\usepackage{array}
\usepackage{tabularx}
\usepackage{adjustbox}
\usepackage{subcaption}
\usepackage{times}
\usepackage{latexsym}

\usepackage{xcolor}

\usepackage{amssymb}  
\usepackage{amsmath}

\usepackage[T1]{fontenc}

\usepackage[utf8]{inputenc}

\usepackage{microtype}
\usepackage{amsmath}
\usepackage{float}
\usepackage{inconsolata}

\usepackage{graphicx}
\usepackage{multicol}
\usepackage{longtable}
\usepackage{multirow}
\usepackage{booktabs} 
\usepackage{array} 
\usepackage{colortbl}
\usepackage[most]{tcolorbox}

\usepackage{comment}

\title{Number Representations in LLMs:\\ A Computational Parallel to Human Perception}

\author{
  \normalsize \textbf{H. V. AlquBoj}$^{*}$ \hspace{5mm} \textbf{Hilal AlQuabeh}$^{*1}$ \hspace{5mm} \textbf{Velibor Bojkovic}$^{*1}$   \\ 
    \normalsize \textbf{Tatsuya Hiraoka}$^{1}$  \hspace{5mm} \textbf{Ahmed Oumar El-Shangiti}$^{1}$ \hspace{5mm} \textbf{Munachiso Nwadike}$^{1}$ 
    \\ \normalsize \textbf{Kentaro Inui}$^{1, 2, 3}$ \\[10pt]
  \normalsize $^{1}$ Mohamed bin Zayed University of Artificial Intelligence (MBZUAI) \\
  \normalsize $^{2}$ Tohoku University, \hspace{5mm} $^{3}$ RIKEN \\
  \normalsize $^*$Amalgamation of first authors' names.
}

\begin{document}
\maketitle
\begin{abstract}

Humans are believed to perceive numbers on a logarithmic mental number line, where smaller values are represented with greater resolution than larger ones. This cognitive bias, supported by neuroscience and behavioral studies, suggests that numerical magnitudes are processed in a sublinear fashion rather than on a uniform linear scale. Inspired by this hypothesis, we investigate whether large language models (LLMs) exhibit a similar logarithmic-like structure in their internal numerical representations. By analyzing how numerical values are encoded across different layers of LLMs, we apply dimensionality reduction techniques such as PCA and PLS followed by geometric regression to uncover latent structures in the learned embeddings. Our findings reveal that the model’s numerical representations exhibit sublinear spacing, with distances between values aligning with a logarithmic scale. This suggests that LLMs, much like humans, may encode numbers in a compressed, non-uniform manner\footnote{Code is available at: \url{https://github.com/halquabeh/llm_natural_log}}\footnote{Correspondence: \{hilal.alquabeh, velibor.bojkovic, kentaro.inui\}@mbzuai.ac.ae}. 
\end{abstract}


\section{Introduction}
\input{sections/new_intro} 
\section{Related Works}
\input{sections/related_work} 
\section{Methodology}
\input{sections/background}

\input{sections/proposal}



\section{Conclusion}
\input{sections/conclusion}



\bibliography{custom}
\newpage

\appendix

\input{sections/appendix}

\end{document}

%% file: sections/new_intro.tex
Large language models (LLMs) have demonstrated impressive capabilities in natural language processing tasks \cite{touvron2023llama1,achiam2023gpt}, yet their internal representations of abstract concepts, i.e., numbers, space, and time, remain largely opaque. Recent research suggests that LLMs construct structured "world models," encoding relationships in ways that can be systematically analyzed \cite{petroni2019language,radford2019language}. For instance, studies have shown that spatial and geographical information is embedded in low-dimensional subspaces, where model performance correlates with data exposure \cite{gurneelanguage,godey2024scaling}. Similarly, numerical representation is influenced by tokenization strategies, with base-10 encoding proving more efficient for numeric reasoning tasks than higher-base tokenizations \cite{zhou2024scaling}.

The \textit{linear hypothesis} of internal representations \cite{park2023linear} posits that concepts in LLMs are structured within geometric, linear subspaces, facilitating interpretability and manipulation. This framework suggests that numerical properties follow systematic, monotonic trends \cite{heinzerling2024monotonic}. As a result, it has been commonly assumed that numerical values are represented in a uniform linear fashion \cite{zhu2025language}. However, recent probing studies \cite{zhu2025language,levy2024language} challenge this assumption, revealing a non-uniform encoding of numbers in LLMs, where precision decreases for larger values. These findings raise questions about how artificial systems internalize numerical representations, particularly in relation to the scaling of numbers. Do LLMs preserve a uniform spacing of numerical values, and if not, what is the nature of their positioning?


\begin{figure}[t]
    \centering
    \includegraphics[width=\linewidth]{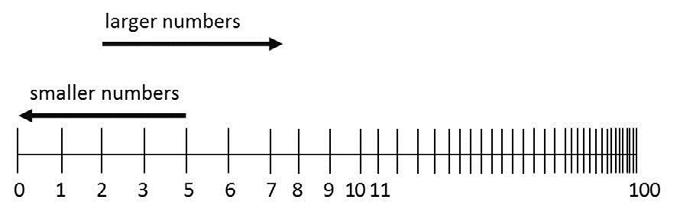} 
    \caption{\textit{Logarthmic mental number line hypothesis} asserts that humans innately percieve numbers on a logarithmic scale. Image source \cite{fritz2013development}.}
    \label{fig log line}
\end{figure}

Such questions naturally lead to an investigation of whether LLMs encode numerical values in a way that mirrors human cognition, as suggested by the \textit{logarithmic mental number line hypothesis}. This hypothesis posits that humans perceive numerical magnitudes nonlinearly, following a logarithmic rather than a uniform linear scale (see Figure \ref{fig log line}). Rooted in psychophysical studies like the Fechner-Weber law, this idea is supported by behavioral experiments showing that young children and individuals with limited formal education tend to map numbers logarithmically when placing them on a spatial axis \cite{fechner1860elemente, dehaene2003neural, siegler2003development}. While formal training shifts numerical perception toward a more linear scale, logarithmic encoding persists in tasks involving estimation and large-number processing \cite{dehaene2008log, moeller2009children}.

Inspired by this, we investigate whether LLMs encode numerical values in a manner analogous to the human logarithmic mental number line. By analyzing hidden representations across model layers, we examine the geometric structure of numerical magnitudes and their underlying trends. Our approach first employs dimensionality reduction techniques, including Principal Component Analysis (PCA) and Partial Least Squares (PLS), to transform the hidden representations onto a one-dimensional number line, that best fits its dominant numerical features. Second, using Spearman rank coefficient and geometric non-linear regression, we specifically test whether two key properties reminiscent of human numerical cognition (order preservation in representations and a compression effect where distances between consecutive numbers decrease as values increase) emerge in LLMs.

While both PCA and PLS reveal that numerical representations largely reside in a linear subspace, only PCA captures systematic sublinearity, suggesting that simple linear probes\footnote{PLS is a linear probe that projects input data onto a lower-dimensional subspace, maximizing covariance with the target.} may overlook the underlying non-uniformity in LLMs’ numerical encoding.



\paragraph{Contributions} We summarize our main finding in the following: 
\begin{itemize}
    \item
    We introduce a methodology for analyzing the geometric structure of number representations, offering a systematic approach to studying numerical abstractions in artificial neural networks.
    \item 
    We provide empirical evidence that LLMs encode numerical values in a structured yet non-uniform way, revealing systematic compression reminiscent of the human logarithmic mental number line. Our findings refine the linear hypothesis by showing that numerical magnitudes in LLMs are not evenly spaced but follow a structured compression pattern.    
\end{itemize}

\begin{figure*}[ht!]
    \centering
    \includegraphics[width=\linewidth]{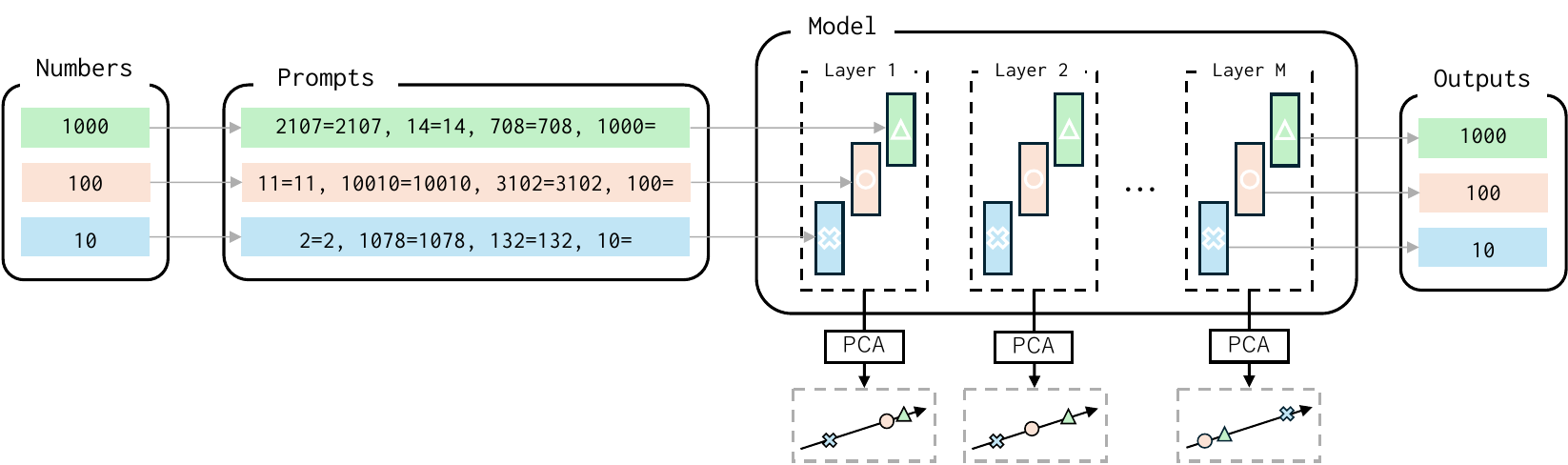} 
    \caption{The overall graphical representation of our method. Numbers are passed to the model in form of a prompt and the internal representations are captured from the embeddings corresponding to token '='. At every layer, we perform PCA projections onto one and two dimensional subspaces and pick a layer with highest explained variance ($\sigma^2$) score to further analyze monotonicity and scaling of number representations.}
    \label{fig pipeline}
\end{figure*}

%% file: sections/related_work.tex


Linearity of internal representations \cite{park2023linear} has been a central assumption in existing research, suggesting that language models encode numerical values in a linear manner. However, \citet{zhu2025language} present a more nuanced perspective. Their analysis of partial number encoding (Appendix F) shows that probing accuracy declines as sequence length increases, with greater difficulty in capturing precise values at larger scales, a pattern reminiscent of logarithmic encoding, where resolution is higher for smaller numbers. Some of the conclusions in \citet{zhu2025language} are that LLMs encode numerical values in their hidden representations, yet linear probes fail to precisely reconstruct these values, as discussed in \citet[Section 3.1]{zhu2025language}. The authors there suggest that ``This phenomenon may indicate that language models use stronger non-linear encoding systems''.  Our findings support this claim and further uncover the underlying nature of this non-linearity.

Recent studies such as \citet{levy2024language, zhou2024scaling} show that LLMs rely on base-10 digit-wise representations rather than encoding numbers in a continuous linear space, as revealed through circular probing techniques. While individual digits are accurately reconstructed, performance declines for larger numbers, suggesting a structured rather than holistic encoding. Furthermore, \citet{zhou2024scaling} demonstrate that LLMs trained on higher-base numeral systems struggle with numerical extrapolation, implying an implicit compressed representation where smaller values have finer granularity—consistent with logarithmic scaling. Collectively, these results align with and further substantiate the hypothesis that LLMs internally represent numbers in a non-uniform, sublinear manner.

Logarithmic functions can appear linear over small local intervals, which may explain why LLMs are often assumed to represent numbers linearly. Consequently, methods like PLS regression and activation patching \cite{heinzerling2024monotonic,elshangiti2024geometrynumericalreasoninglanguage}, which analyze small activation variations, may capture local monotonicity while missing the global nonlinear structure. This suggests that reported linear effects could stem from analyzing narrow numerical ranges, whereas a broader examination may reveal an underlying logarithmic representation. 

Finally, we also emphasize the difference between our work and prior studies on numerical reasoning \cite{park2022measurements, zhang2020language} which evaluate models' ability to process explicit numbers rather than probing their internal representations. While \citet{park2022measurements} focus on tasks like unit conversion and range detection, \citet{zhang2020language} examine numerical magnitude in common sense reasoning. Unlike these works, our study investigates the spatial structure of numerical representations within hidden states and how this encoding generalizes across scales.

%% file: sections/background.tex



\subsection{General settup} 

The \textit{logarithmic mental number line hypothesis} \cite{dehaene2008log} suggests that humans innately perceive numerical magnitudes on a logarithmic scale rather than a linear one. Formally, if we denote the internal mapping of numbers to their cognitive representation as \( f_{\text{H}} \), the hypothesis asserts that \( f_{\text{H}} \) is approximately logarithmic. While the exact nature of \( f_{\text{H}} \) remains elusive, we adopt the guiding principle \( f_{\text{H}} \equiv \log \) in our experiments.

On the other side, Large language models, like LLaMA-2, process inputs by mapping them into a high-dimensional representation space, where each input \( x \) (e.g., a number) is transformed into an internal representation \( f(x) \in \mathbb{R}^d \). Analyzing the geometry of these representations across a set of inputs \( \mathcal{X} \) can reveal how the model organizes and reasons about them, shedding light on emergent properties such as whether the model encodes numerical values along a number line or exhibits cognitive-like patterns, such as sublinear scaling. 

Our goal is \textbf{to study the properties of the function $f_{\text{LLM}}$, which serves as a counterpart of the human cognitive mapping $f_{\text{H}}$ described above}. Specifically, we examine  whether $f_{\text{LLM}}$ preserves the natural ordering of numbers and how it transforms their magnitudes. 



\subsection{Definition of $f_{\text{LLM}}$}
To analyze the structure of hidden representations, we apply a projection \( T: \mathbb{R}^d \to \mathbb{R}^p \), $p=1,2$, obtained from techniques such as Principal Component Analysis (PCA) or Partial Least Squares (PLS). Then, our function is given by
\begin{equation}\label{eq projection p_T}
        f_{\text{LLM}}(x) := T(f(x)),
\end{equation}
where $f$ is a map between the input number and the corresponding internal representation in the model (see Sections \ref{experiment1} and \ref{experimen2} how $f$ is defined in various settings).

For any two inputs \( x, y \in \mathcal{X} \), we define the distance between their projections following the mapping $T$ using Euclidean norm as  
\begin{equation}
d(x, y) = ||f_{\text{LLM}}(x) - f_{\text{LLM}}(y)||.
\end{equation}


\subsection{Abstract Number Line and Monotonicity Metric}  
If the projection dimension is $p=1$, one-dimensional embedding of numerical inputs forms a \emph{number line} if the projections preserve monotonicity (resp. reverse monotonicity), i.e., for \( x_1 < x_2 \) (resp. $x_1>x_2$), we have  
$
f_{\text{LLM}}(x_1) < f_{\text{LLM}}(x_2).
$
This ensures that the natural order of numerical values is maintained in the representation space.  



    
    
  

To measure monotonicity properties of the function $f_{\text{LLM}}$ we use Spearman rank correlation that we briefly describe next. Let $X, Y\in \mathbb{R}^n$ be two real $n$-dimensional vectors and let $R(X)$ (resp. $R(Y))$ denote an $n$-dimensional vector obtained from $X$ (resp. $Y$) where the entries are substituted with their ranks in the sequence of sorted entries of $X$ (resp. $Y$). Then, Spearman rank correlation coefficients (usually denoted by $\rho$) is given by:
\begin{equation}\label{eq spearman general}
    \rho=\frac{\textbf{Cov}(R(X),R(Y))}{\sigma(R(x))\cdot\sigma(R(Y))},
\end{equation}
where $\textbf{Cov}(R(X),R(Y))$ is the covariance between rank vectors $R(X)$ and $R(Y)$, while $\sigma(R(X))$ and $\sigma(R(Y))$ are their respective standard deviations. 

Spearman coefficient $\rho$ is a nonparametric measure for the alignment of the two vectors. Loosely speaking, the coefficient assesses if the increment in one variable corresponds to the increase (or decrease) of the other. In particular, unlike Pearson coefficient which takes into account the value of the changes, Spearman's $\rho$ takes into account only the sign of the changes. 



\subsection{Scaling Rate Index}

For a monotonically increasing sequence of positive real numbers $x_1,\dots, x_n$, we introduce the Scaling Rate Index to measure the rate at which numbers grow in magnitude. More precisely, we seek for positive real constants $\alpha$ and $\beta$ that minimize the following objective function:
\begin{equation}\label{eq sri}
    \sum_{i=1}^{n-1}|(x_{i+1}-x_i) - \alpha\cdot \beta^i|^2.
\end{equation}


In particular, if $\beta>1$, the difference between two consecutive members $x_i$ and $x_{i+1}$ is expected to increase with $i$. In other words, $x_{i+1}-x_{i}$ is expected to be smaller than $x_{i+2}-x_{i+1}$. Such sequences are commonly known as \textit{convex sequences} \cite{rockafellar2015convex} and the growth of $x_i$ is exponential (superlinear) in $i$. An example of such a sequence that is relevant to our study is that defined with $x_i:=10^i$. Then, $x_{i+1}-x_i= 9\cdot 10^{i}$ and we can take $\alpha=9$ and $\beta = 10$. 

If $\beta=1$, expression \eqref{eq sri} indicates that the difference between two consecutive members $x_i$ and $x_{i+1}$ is approximately constant, i.e. the sequence is approximately linearly increasing. For example, one may take the sequence $x_i:=i$, for which $\alpha=1=\beta$.

Finally, $\beta<1$ implies that the difference between consecutive members is steadily decreasing, $x_{i+1}-x_{i}$ is expected to be greater than $x_{i+2}-x_{i+1}$. Such sequences are commonly known as \textit{concave sequences} \cite{rockafellar2015convex}, and the growth of the sequence is exponentially decreasing (sublinear). An example of such a sequence is given by $x_i = 1- \frac{1}{10^i}$, with $\alpha = 9$ and $\beta = \frac{1}{10}$.

%% file: sections/proposal.tex
\section{Experiment 1: Identifying number line using contextualized numbers}\label{sec exp 1}
\label{experiment1}
\paragraph{Setup.} 
To systematically probe the model’s numerical representations, we partition numbers into logarithmically spaced groups:
\begin{equation}\label{eq groups G}
\begin{aligned}
    &G_1 = \{ 1, 2, \dots, 20\}, \\ &G_{i+1} = \{10^i - 19, \dots, 10^i + 20\}, \quad i \geq 2.
\end{aligned}
\end{equation}
These groups ensure that larger group indices correspond to numerical magnitudes that increase exponentially with index, reflecting the logarithmic nature of the mental number line hypothesis.

To analyze the embeddings of numbers, to every number \( x \in G_i \) we assign the following prompt:
\begin{equation}\label{eqn:prompt}
    x \leftarrow \text{\texttt{a=a, b=b, c=c, x=}}
\end{equation}

\noindent where \( \text{\texttt{a, b, and c}} \) are randomly generated numbers from the groups $G_i$. This prompt structure is designed to provide the model with contextual examples, encouraging it to invoke the number \( x \) in model's hidden states representations (see Figure \ref{fig pipeline}). Such approaches have been used in prior work to probe contextual representations in language models \citet{srivastava-etal-2024-nice}.

 The hidden state representation \( f(x) \in \mathbb{R}^d \) is extracted from a designated layer of the model from the last token in the prompt, e.g. the `=` token. We use only those $x$ for which the generated output of the model is $x$ itself.

To ensure a representative sampling, we randomly select \( k \) numbers from each group \( G_i \). These sampled numbers collectively form a dataset, denoted as \( \mathcal{X} \), which serves as the basis for our analysis. The set of hidden state representations, \( \{f(x)\}_{x \in \mathcal{X}} \), is then aggregated and analyzed to investigate patterns and properties in the embedding space.

To control for potential biases introduced by tokenization (where larger numbers often span more tokens) we conduct a complementary experiment using non-numerical sequences. Instead of numerical inputs, we construct sequences of random letters with lengths corresponding to the tokenized representations of numbers. The letter sequences are grouped to their lengths so that the grouping approximately matches one of the numbers, and the prompts corresponding to specific letter sequences are designed in a similar fashion as for the numbers (\ref{eqn:prompt}). This setting allows us to compare any observed structural patterns between the number representations and letter representations. By doing so, we can determine whether the model truly encodes numerical magnitude or if it is simply responding to surface-level features of the input.


\paragraph{Motivation.} 
The goal of this experiment is twofold. We first investigate whether LLMs encode numerical values in context along a \textit{monotonic number line} in their internal representation space. Second, we test whether this number line exhibits \textit{sublinear scaling}, similar to human cognitive representations of numbers.


\paragraph{Methodology.} 
For these purposes, we use:
\vspace{0.5\baselineskip}

\noindent $\bullet{}$ PCA nad PLS. After the set of hidden state representations is aggregated, we further project it into a one-dimensional space using PCA or PLS methods. In particular, for PLS we take the numbers themselves to form the target vectors, while for letters, we consider the letter sequence as a base $26$ representation of a number (with random but fixed assignment of values to the letters), and use this number as the corresponding target.

\vspace{0.5\baselineskip}
\noindent $\bullet{}$ Monotonicity metric. We apply it on a sequence $x_1,\dots,x_n$ of all the numbers in the union of groups $G_j$ defined in \eqref{eq groups G}, and their respective projections $f_{\text{LLM}}(x_1),\dots,f_{\text{LLM}}(x_n)$. Spearman rank coefficient will tell us whether the model preserves natural ordering of the numbers.



\vspace{0.5\baselineskip} 
\noindent $\bullet{}$ Scaling Rate Index. The initial centers of \( G_i \), given by \( x_i = 10^i \), form a convex, exponentially growing sequence characterized by parameters \( \alpha = 9 \) and \( \beta = 10 \). These numbers serve as representative scales of the numbers within each group.

To obtain a robust estimate of how these scales are preserved in the projections under $f_{\text{LLM}}$, we compute the expectation of projections in each group. Specifically, for SRI analysis, we define the sequence

\begin{equation}\label{eq expected center}
\bar{f}_{\text{LLM}}(i) = \underset{{x \in G_i}}{\mathbb{E}}[f_{\text{LLM}}(x)],
\end{equation}

\noindent and fit positive constants $\alpha$ and $\beta$ such that $\bar{f}_{\text{LLM}}(i)\approx \alpha\cdot\beta^i$. 

In particular, the mapping 
\begin{equation}
10^i \mapsto \bar{f}_{\text{LLM}}(i) 
\end{equation}
allows us to examine how does $f_{\text{LLM}}$ scales numerical magnitudes. The fundamental question we seek to answer is: \textbf{What is the nature of the function \( f_{\text{LLM}} \) (logarithmic, linear, or exponential)}? 




To answer this question, we analyze the scaling factor $\beta$ in the fitted exponential model\footnote{We can disregard $\alpha$ from the analysis since it does not influence the scaling but merely introduces a bias.}. In the following, we explain how different values of $\beta$ correspond to the underlying properties of $f_{\text{LLM}}$.

    $\bullet{}$ If \( \beta > 1 \), the sequence \( \bar{f}_{\text{LLM}}(i) \) is convex and exponentially increasing. This means \( f_{\text{LLM}} \) maps an exponentially increasing sequence to another exponentially increasing sequence:
    \[
    9 \cdot 10^i \mapsto \alpha \cdot \beta^i = \alpha \cdot 10^{(\log_{10} \beta) \cdot i}.
    \]
    Thus, \( f_{\text{LLM}} \) preserves the original spacing of numbers, albeit with a scaling factor \( \log_{10} \beta >0 \).  

    $\bullet{}$ If \( \beta = 1 \), the sequence \( \bar{f}_{\text{LLM}}(i) \) is linearly increasing, meaning \( f_{\text{LLM}} \) takes the form:
    \[
    10^i \mapsto \alpha \cdot i = \alpha \cdot \log_{10} 10^i.
    \]
    In this case, \( f_{\text{LLM}} \) exhibits logarithmic scaling.  

    $\bullet{}$ If \( \beta < 1 \), the sequence \( \bar{f}_{\text{LLM}}(i) \) is concave, exponentially decaying. Here, \( f_{\text{LLM}} \) follows:
    \[
    10^i \mapsto \alpha \cdot \beta^i = \alpha \cdot 10^{-(\log_{10} \frac{1}{\beta}) \cdot i}.
    \]
    Thus \( f_{\text{LLM}} \) is a \textit{sub-logarithmic} mapping.

\begin{table}[h]
\begin{footnotesize}
    \centering
    \resizebox{.48\textwidth}{!}{ 
    \renewcommand{\arraystretch}{1.2}
    \setlength{\tabcolsep}{3pt}
    \begin{tabular}{lccccc}
        \toprule
        \textbf{Model} & \textbf{Group} & \textbf{Layer} & $\rho \pm$ std & $\beta \pm$ std & $\sigma^2$ $\pm$ std \\
        \midrule
        \multirow{2}{*}{LLaMA-2-7B} & Numbers & 3 & 0.97 $\pm$ 0.00 & 0.83 $\pm$ 0.06 & 0.60 $\pm$ 0.01 \\
         & Letters & 1 & 0.45 $\pm$ 0.00 & 1.21 $\pm$ 0.00 & 0.24 $\pm$ 0.00 \\
        \midrule
        \multirow{2}{*}{Pythia-2.8B} & Numbers & 8 & 0.94 $\pm$ 0.01 & 0.54 $\pm$ 0.01 & 0.31 $\pm$ 0.01 \\
         & Letters & 11 & 0.89 $\pm$ 0.01 & 0.53 $\pm$ 0.10 & 0.16 $\pm$ 0.01 \\
        \midrule
        \multirow{2}{*}{GPT-2-L} & Numbers & 18 & 0.95 $\pm$ 0.00 & 0.58 $\pm$ 0.02 & 0.32 $\pm$ 0.00 \\
         & Letters & 5 & 0.11 $\pm$ 0.05 & 0.80 $\pm$ 0.42 & 0.21 $\pm$ 0.01 \\
        \midrule
        \multirow{2}{*}{Mistral-7B} & Numbers & 3 & 0.96 $\pm$ 0.00 & 1.05 $\pm$ 0.00 & 0.44 $\pm$ 0.00 \\
         & Letters & 14 & 0.89 $\pm$ 0.00 & 0.60 $\pm$ 0.00 & 0.22 $\pm$ 0.00 \\
        \midrule
        \multirow{2}{*}{LLaMA-3.1-8B} & Numbers & 1 & 0.41 $\pm$ 0.04 & 1.14 $\pm$ 0.05 & 0.48 $\pm$ 0.01 \\
         & Letters & 1 & 0.56 $\pm$ 0.00 & 0.16 $\pm$ 0.07 & 0.19 $\pm$ 0.01 \\
        \midrule
        \multirow{2}{*}{LLaMA-3.2-Instruct-1B} & Numbers & 4 & 0.93 $\pm$ 0.02 & 1.33 $\pm$ 0.12 & 0.35 $\pm$ 0.01 \\
         & Letters & 1 & 0.57 $\pm$ 0.06 & 0.47 $\pm$ 0.08 & 0.17 $\pm$ 0.00 \\
        \bottomrule
    \end{tabular}
    }
    \end{footnotesize}
    \caption{Comparison of several models on Numbers and Letters groups, evaluated using three metrics: $\rho$, $\beta$, and Explained Variance ($\sigma^2$). Results are reported for the layer with the highest $\sigma^2$ score. Standard deviations are included.}
    \label{tab:model_comparison_compact}
\end{table}

\begin{table}[h]
\begin{footnotesize}
    \centering
    \resizebox{.48\textwidth}{!}{ 
    \renewcommand{\arraystretch}{1.2}
    \setlength{\tabcolsep}{3pt}
    \begin{tabular}{lccccc}
        \toprule
        \textbf{Model} & \textbf{Group} & \textbf{Layer} & $\rho \pm$ std & $\beta \pm$ std & $R^2 \pm$ std \\
        \midrule
        \multirow{2}{*}{Llama-3.2-1B-Instruct} & Numbers & 6 & 0.91 $\pm$ 0.00 & 1.93 $\pm$ 0.05 & 0.68 $\pm$ 0.01 \\
         & Letters & 10 & 0.93 $\pm$ 0.00 & 0.97 $\pm$ 0.03 & 0.45 $\pm$ 0.03 \\
        \midrule
        \multirow{2}{*}{Pythia-2.8b} & Numbers & 1 & 0.78 $\pm$ 0.02 & 4.65 $\pm$ 1.32 & 0.71 $\pm$ 0.01 \\
         & Letters & 20 & 0.90 $\pm$ 0.01 & 0.95 $\pm$ 0.11 & 0.46 $\pm$ 0.04 \\
        \midrule
        \multirow{2}{*}{GPT2-L} & Numbers & 17 & 0.96 $\pm$ 0.01 & 1.15 $\pm$ 0.09 & 0.67 $\pm$ 0.03 \\
         & Letters & 33 & 0.81 $\pm$ 0.03 & 0.93 $\pm$ 0.04 & 0.44 $\pm$ 0.01 \\
        \midrule
        \multirow{2}{*}{Llama-2-7b} & Numbers & 5 & 0.93 $\pm$ 0.00 & 2.62 $\pm$ 0.00 & 0.81 $\pm$ 0.00 \\
         & Letters & 27 & 0.88 $\pm$ 0.03 & 0.91 $\pm$ 0.02 & 0.45 $\pm$ 0.01 \\
        \midrule
        \multirow{2}{*}{Mistral-7B-v0.1} & Numbers & 7 & 0.88 $\pm$ 0.00 & 14.87 $\pm$ 8.28 & 0.81 $\pm$ 0.00 \\
         & Letters & 29 & 0.86 $\pm$ 0.00 & 1.62 $\pm$ 0.00 & 0.63 $\pm$ 0.00 \\
        \midrule
        \multirow{2}{*}{Llama-3.1-8B} & Numbers & 4 & 0.93 $\pm$ 0.01 & 2.00 $\pm$ 0.01 & 0.73 $\pm$ 0.01 \\
         & Letters & 16 & 0.93 $\pm$ 0.01 & 0.88 $\pm$ 0.06 & 0.45 $\pm$ 0.02 \\
        \bottomrule
    \end{tabular}
    }
\end{footnotesize}
\caption{Comparison of several models on Numbers and Letters groups, evaluated using three metrics: $\rho$, $\beta$, and $R^2$. Results are reported for the layer with the highest $R^2$. Standard deviations are included.}
\label{tab:model_comparison}
\end{table}

\begin{figure}[ht!]
    \centering
    \begin{subfigure}[b]{0.23\textwidth}
        \centering
    \includegraphics[width=\textwidth]{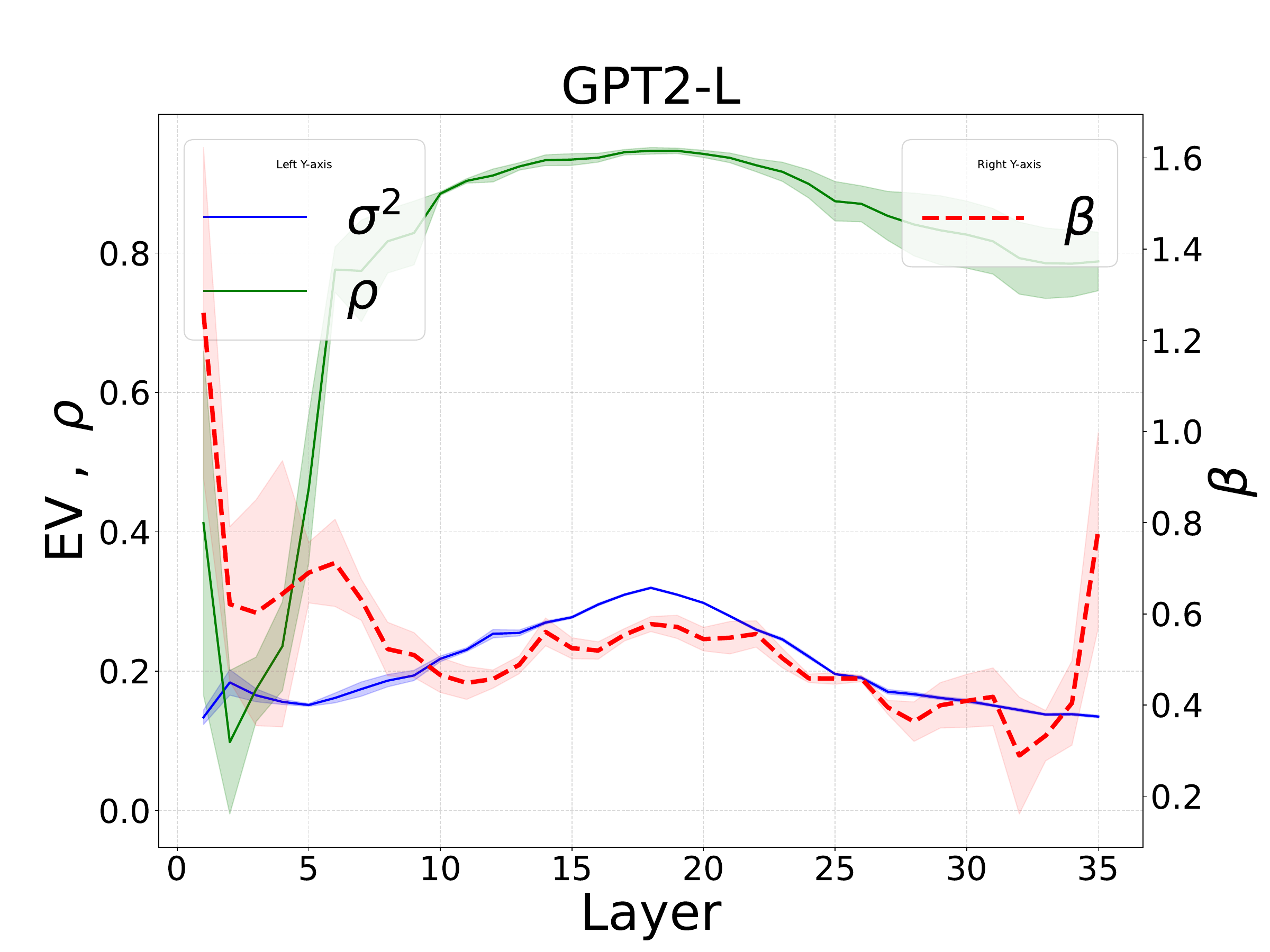}
    \end{subfigure}
    \hfill
    \begin{subfigure}[b]{0.23\textwidth}
        \centering
        \includegraphics[width=\textwidth]{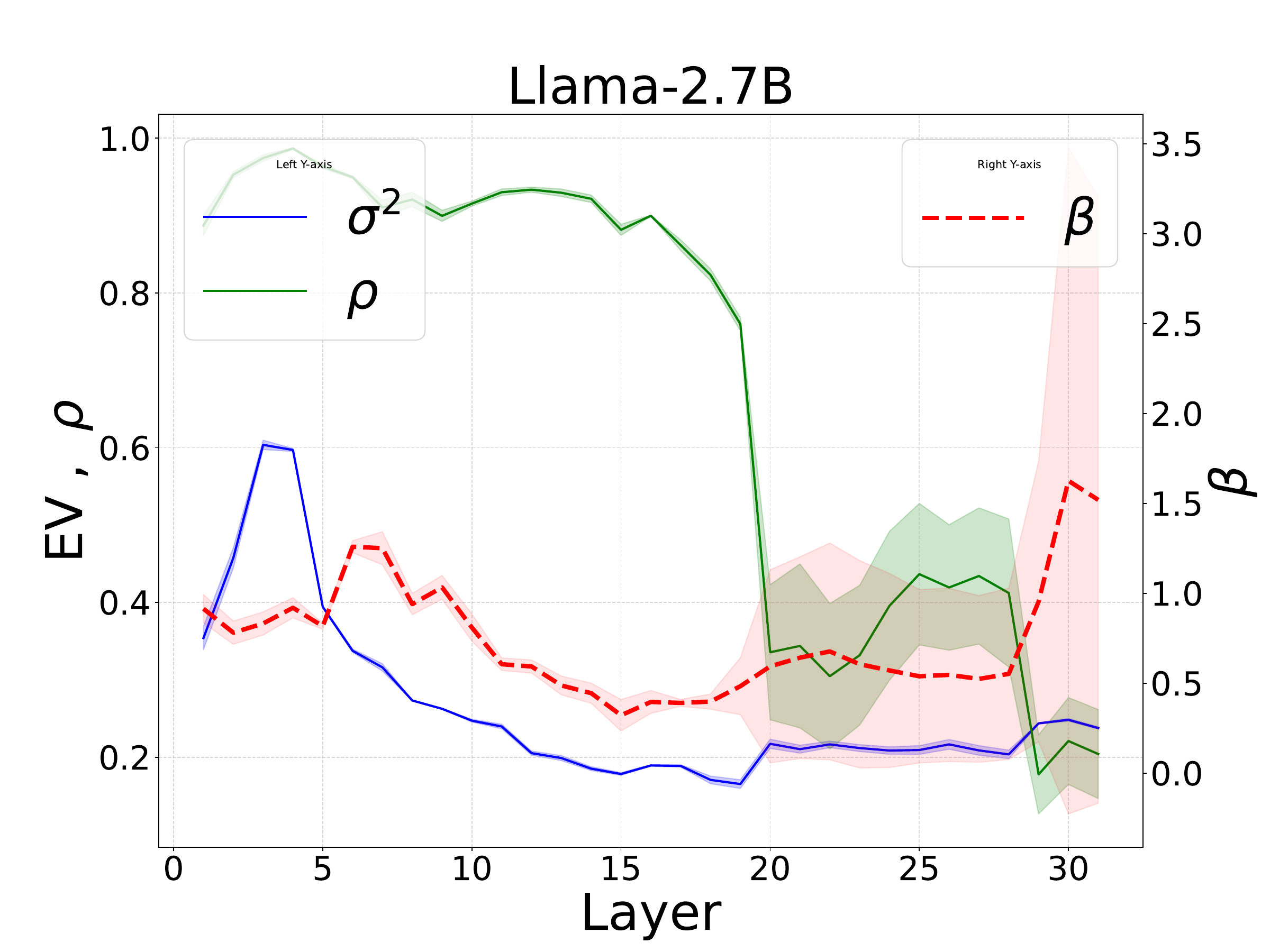}
    \end{subfigure}
    
    \vspace{0.3cm}  
    
    \begin{subfigure}[b]{0.23\textwidth}
        \centering
        \includegraphics[width=\textwidth]{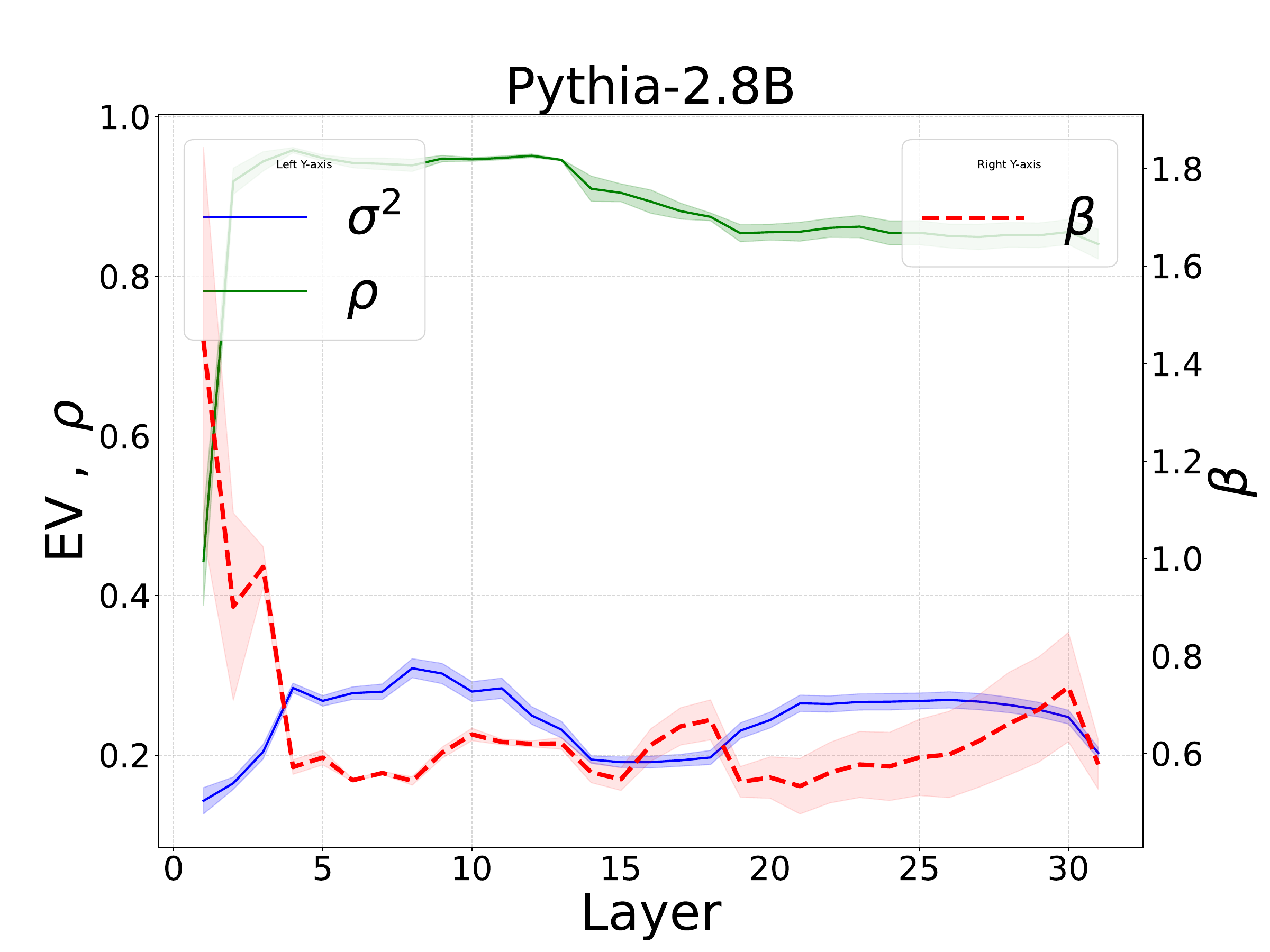}
    \end{subfigure}
    \hfill
    \begin{subfigure}[b]{0.23\textwidth}
        \centering
        \includegraphics[width=\textwidth]{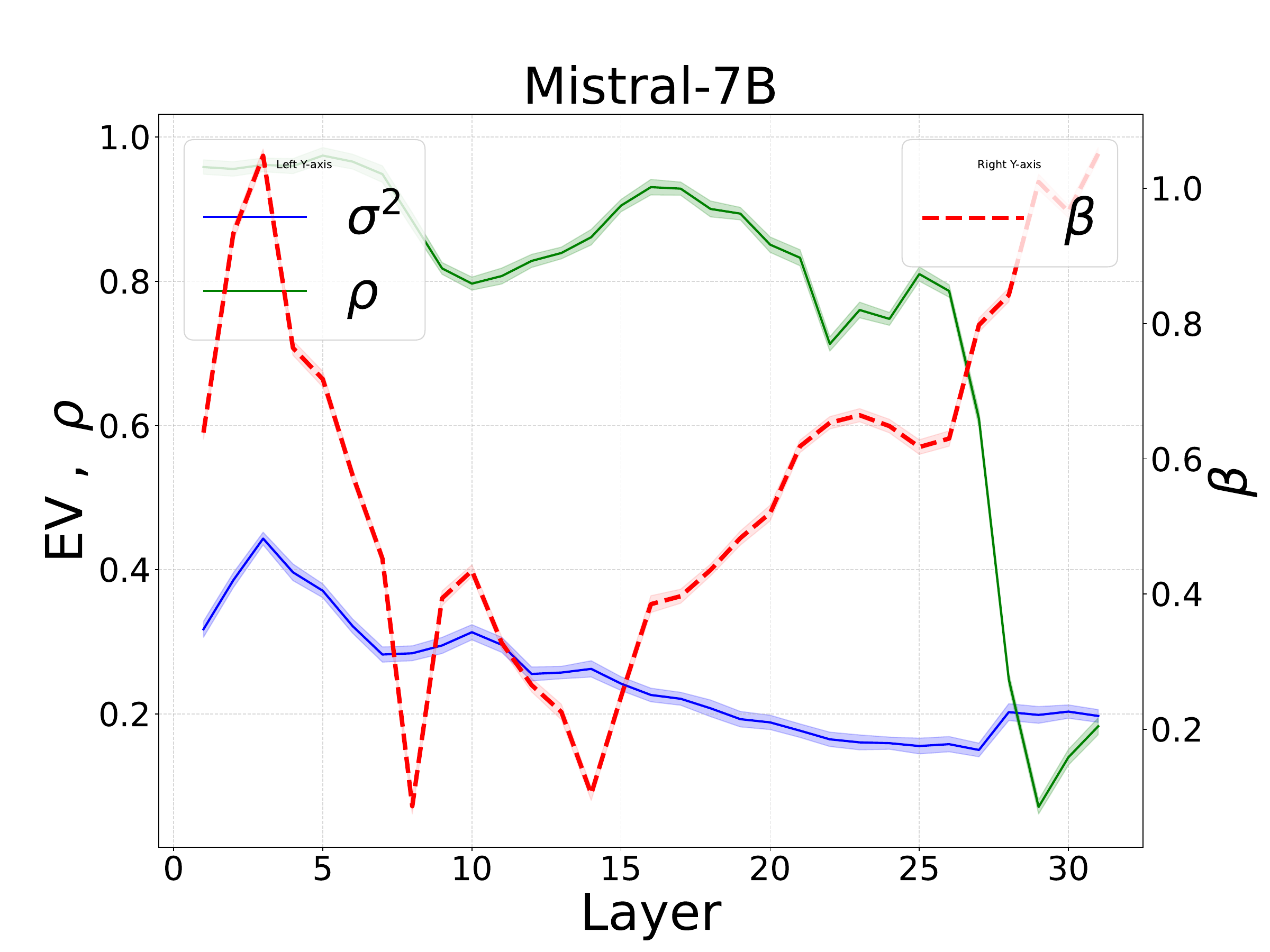}
    \end{subfigure}

\caption{Layer-wise analysis of four models on numerical groups, showing explained variance ($\sigma^2$), monotonicity ($\rho$), and Scaling Rate Index ($\beta$). The layer with maximum $\sigma^2$ aligns with peak $\rho$, indicating optimal numerical encoding.}
    \label{fig:metrics-layers-numerics}
\end{figure}

\begin{figure}[ht]
    \centering
\includegraphics[width=0.9\linewidth]{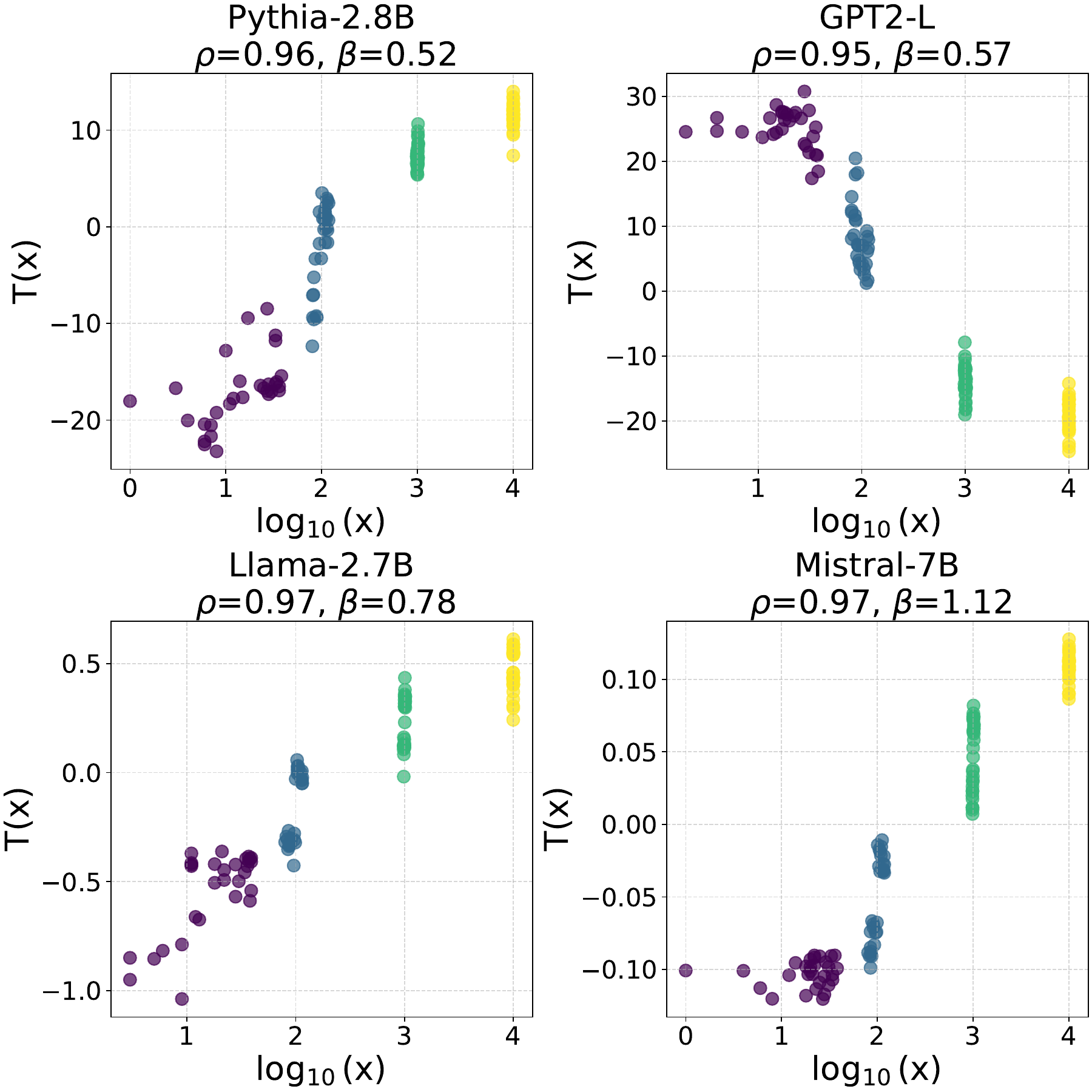}
\caption{Projections of numerical representations (y-axis) against their log-scaled magnitudes (x-axis) for the layer with the highest explained variance in four models. Sublinearity and monotonicity ($\rho$) are indicated above each subfigure, demonstrating consistent sublinear trends and strong monotonic relationships across models.}    \label{fig:projections-numbers}
\end{figure}

\begin{figure}[ht]
    \centering
\includegraphics[width=0.9\linewidth]{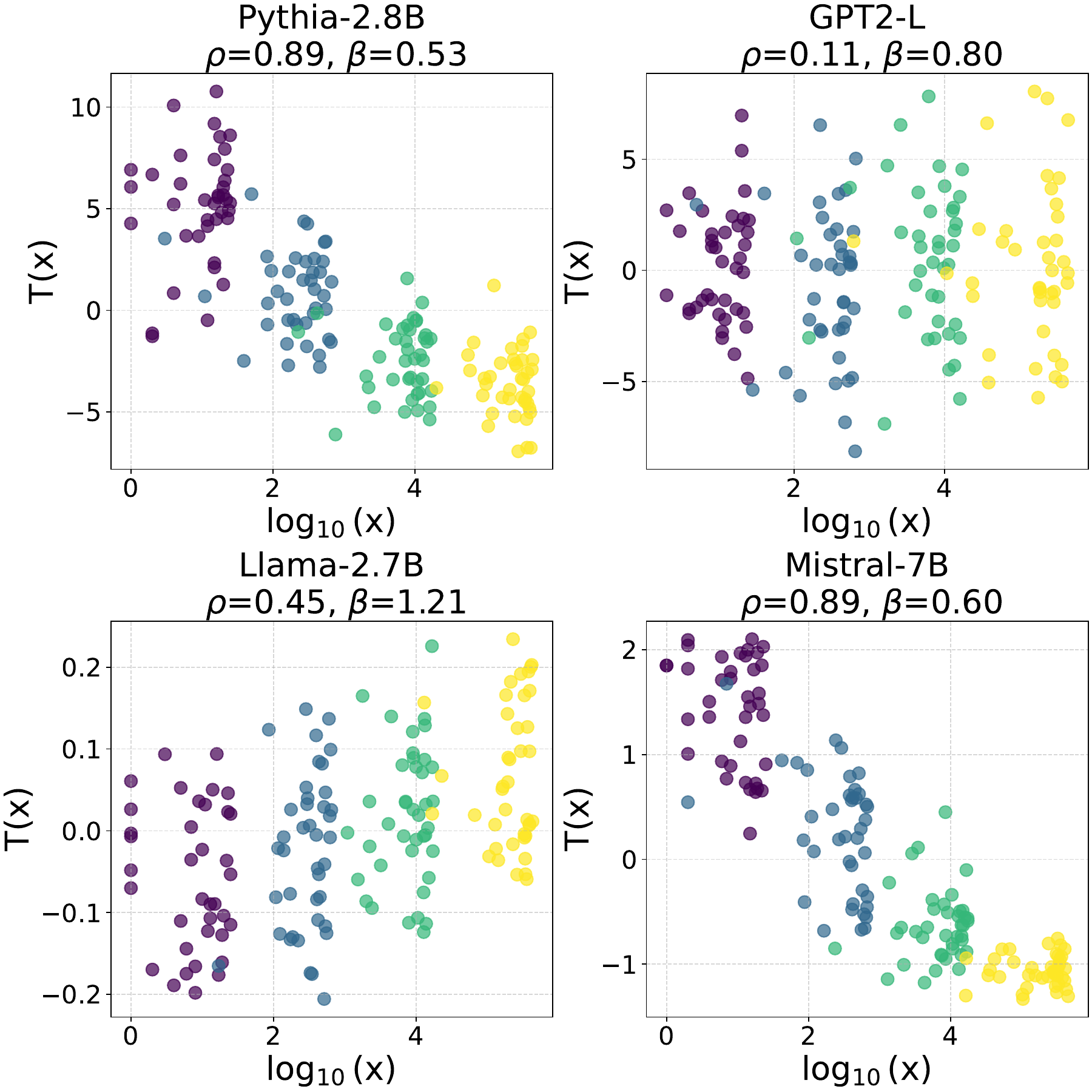}
\caption{Projections of letters representations (y-axis) against their log-scaled magnitudes (x-axis) assigned proportional to their length, for the layer with the highest explained variance in four models. Sublinearity and monotonicity ($\rho$) are indicated above each subfigure, demonstrating consistent sublinear trends and strong monotonic relationships across models.}    \label{fig:projection_letters}
\end{figure}

\paragraph{Results.} 

The results reveal distinct yet consistent patterns in how different models encode numerical and alphabetical structures, with variations across layers (Tables \ref{tab:model_comparison_compact} and \ref{tab:model_comparison}). Despite these variations, similar trends emerge across the models, leading to consistent conclusions about their processing of numerical values (please refer to appendix \ref{app:exp} for experimental details).

\paragraph{Numerical vs. Symbolic Representations.} 
First key finding is that numerical embeddings exhibit a significantly higher explained variance ($\sigma^2$ in Table \ref{tab:model_comparison_compact} and $R^2$ in Table \ref{tab:model_comparison}) in the one-dimensional PCA and PLS transformations compared to letter-based embeddings (refer to Methodology). This suggests that numbers naturally align along a one-dimensional manifold—akin to a number line—while random sequences of letters do not display the same structured behavior.

Furthermore, the monotonicity metric ($\rho$) consistently shows higher values for numerical data, with most models achieving $\rho > 0.9$ in both PCA and PLS analyses. This supports the idea that numerical representations are not only structured, but also maintain a well-ordered progression across layers.

The resulting projections obtained using PCA for the numerical and letters groups are visualized in Figures \ref{fig:projections-numbers} and \ref{fig:projection_letters}, respectively. 

\paragraph{Sublinearity in Numerical Representations.}
The sublinearity coefficient ($\beta$) derived from PCA projections reveals notable differences across models. Some, such as LLaMA-2-7B, Pythia, and GPT-2 Large, exhibit strong sublinear (sublogarithmic) scaling with $\beta < 1$, indicating that embedding distances grow at a diminishing rate. In contrast, models like Mistral show a nearly logarithmic trend ($\beta \approx 1$), while others approach a more linear spacing pattern with higher $\beta$ values.

\paragraph{Layer-Wise Dynamics.}
Since Table 1 reports values from the layer with the highest explained variance, interpretation requires caution—other layers with comparable $\sigma^2$ values may exhibit similar trends. Figure \ref{fig:metrics-layers-numerics} provides a layer-wise analysis for four models, demonstrating how sublinearity evolves across different depths.

\paragraph{PCA vs PLS.}
The PLS method achieves high monotonicity ($\rho$) and explained variance ($R^2$) but exhibits lower sublinearity compared to PCA. This discrepancy arises because PLS operates as a supervised linear probe, where the regression target (e.g., numerical values) directly influences the projection. 
This process distorts the intrinsic spacing between points, as PLS prioritizes maximizing covariance with the target over preserving the original geometric structure. In contrast, PCA, being unsupervised, retains the relative spacing of data points in the latent space, better capturing the underlying sublinear trends. This distinction is evident in tables \ref{tab:model_comparison_compact} and \ref{tab:model_comparison}: PCA consistently reveals stronger sublinearity, while PLS achieves higher $R^2$ and $\rho$ by aligning the projection with the target variable. 

Notably, this aligns with findings in \citet{zhu2025language}, where a linear probe  failed to adequately capture the non-linear scaling of hidden states, particularly for larger numbers, where non-linearity becomes more pronounced. Our work explicitly quantify sublinearity using the Scaling Rate Index (SRI, $\beta$), which directly measures the rate of scaling in the latent space. This allows us to better capture the true geometric organization of numerical representations, especially in regimes where non-linear effects dominate.

\begin{figure}[ht]
    \centering
    \begin{subfigure}[b]{0.48\textwidth}
        \centering
    \includegraphics[width=\textwidth]{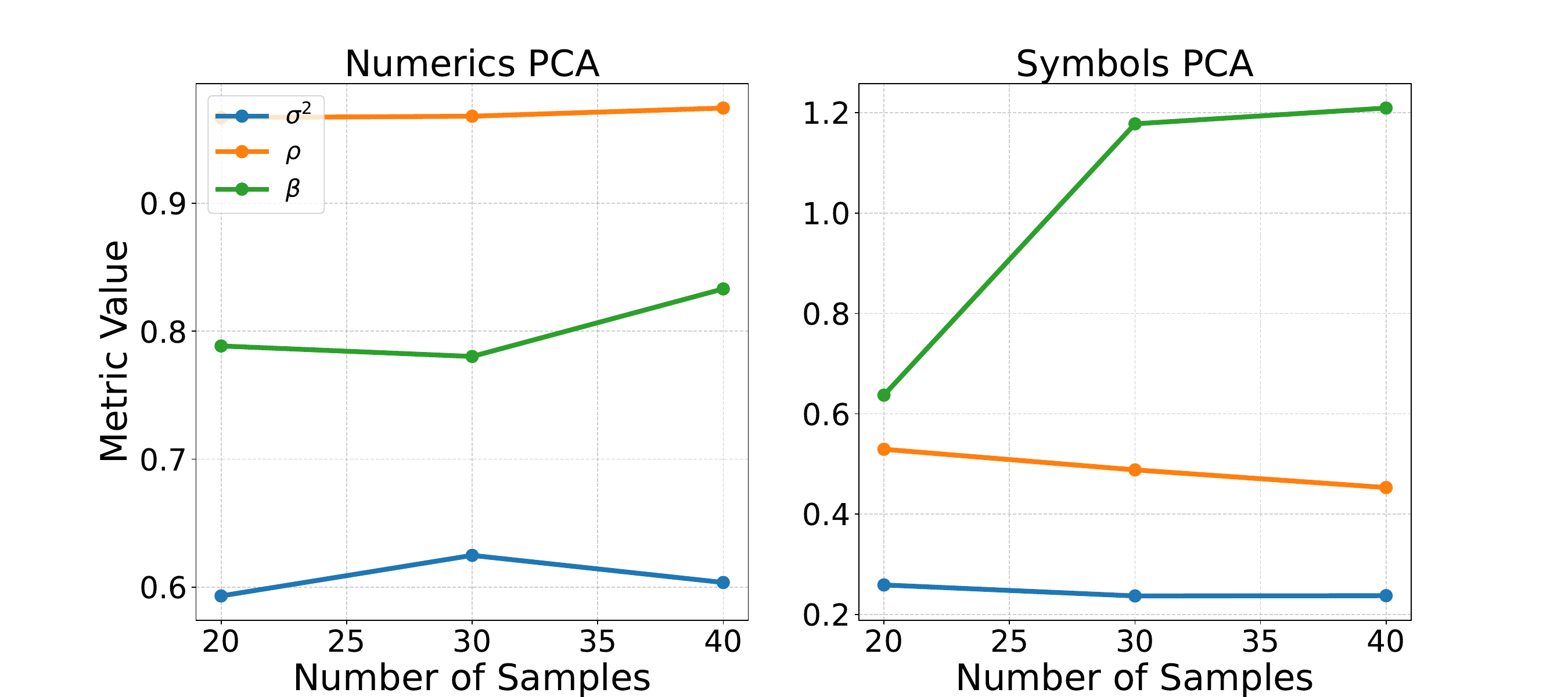}
    \end{subfigure}
    \hfill
    \begin{subfigure}[b]{0.48\textwidth}
        \centering
    \includegraphics[width=\textwidth]{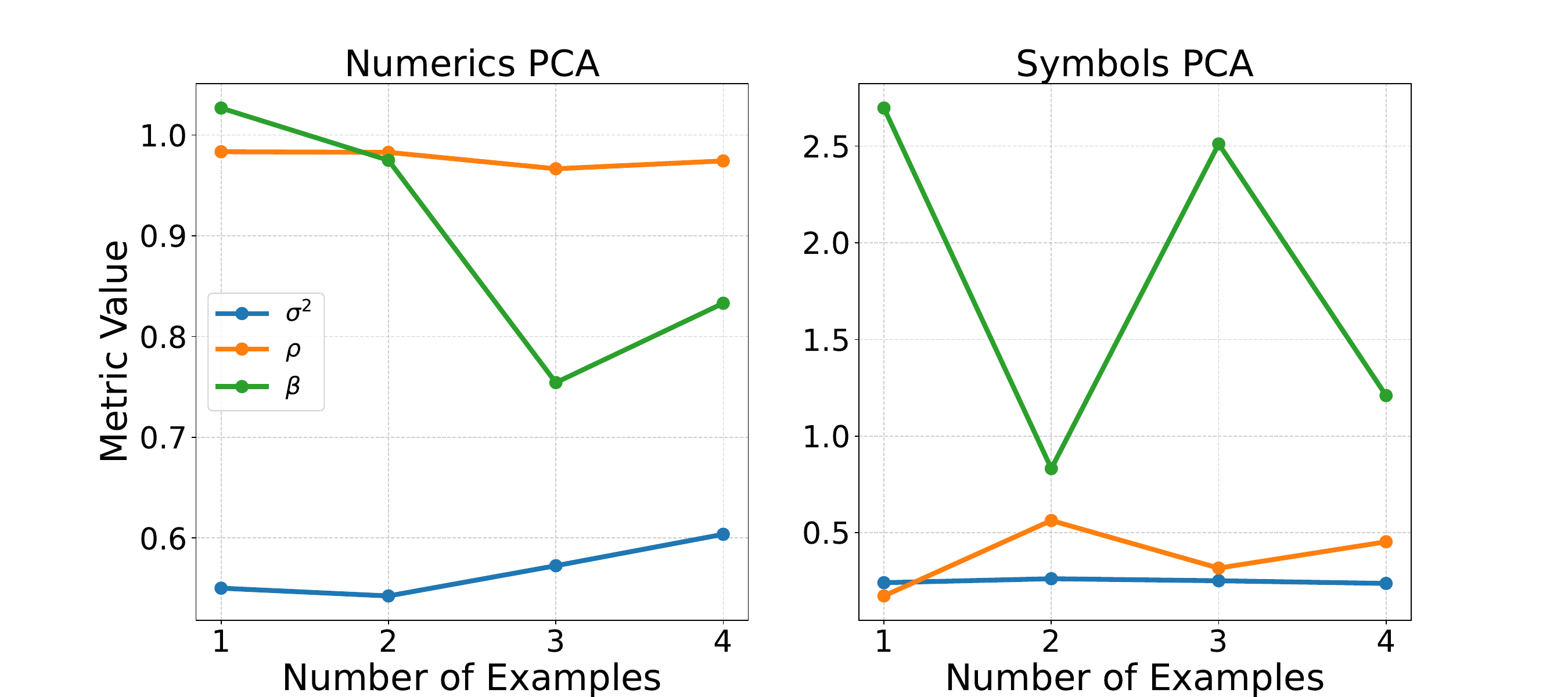}
    \end{subfigure}
    \caption{Top row: Change in metrics with respect to the number of samples in the set $\mathcal{X}$. Bottom row: Change in metrics with respect to the number of in-context examples in the prompt. Left column corresponds to the \textbf{numbers} group, and the right column corresponds to the \textbf{letters} group. Sublinearity and monotonicity trends are highlighted for each case.}
    \label{fig:metrics-ablation}
\end{figure}

\paragraph{Ablation study.} Finally, we perform an ablation study to examine the dependence of our results on the number of examples in the prompt for both numerical and alphabetical datasets. Figure \ref{fig:metrics-ablation} shows that the metrics exhibit greater stability for numerical data compared to alphabetical data, indicating that the model processes numerical information more consistently, while alphabetical representations are more sensitive to prompt variations.

\section{Experiment 2: Identifying number line using real-world tasks}
\label{experimen2}

\paragraph{Setup.} In the previous experiment (Section \ref{experiment1}) we created an artificial experimental setting to test our hypothesis. 
In this experiment, however, we want to further validate our hypothesis using real-world data. We collect names of celebrities along with their birth years and population of different cities/countries from Wikidata~\cite{Wikidata}. The task here is to investigate for similar patterns and observations seen in the previous experiment.

\paragraph{Motivation.}
The goal of this experiment is to investigate how LLMs internally represent numerical values in real-world contexts, specifically focusing on the monotonicity and scaling of these representations. By analyzing the hidden states, we aim to uncover whether the models encode numerical information in a structured and interpretable manner.

\paragraph{Methodology.}

The experimental setup for this experiment is as follows:

$\bullet{}$ \textbf{Prompting the Model:} We prompt the model to provide the exact birth year or population size for each entity in our dataset, which consists of 1K samples.An example of a prompt would be ``What is the population of [country]?''
$\bullet{}$ \textbf{Collecting Model Outputs:} We collect the LLM's output answers, and filter out non-numerical and incorrect responses. 

$\bullet{}$ \textbf{Extracting Hidden States:} We extract the hidden state corresponding to the question mark token at each model layer\footnote{We divided the hidden states into four equally sized groups, ranging from the minimum to the maximum answers, to facilitate the calculation of the Scaling Rate Index (SRI).}.

$\bullet{}$ \textbf{Training PLS Models:} We train one- and two-component PLS models on the extracted hidden states to predict the birth years or population sizes of the entities. This is performed for two LLMs: \textit{Llama-3.1-8B} and \textit{Llama-3.2-1B}.


\paragraph{Results.}

The results, as shown in table \ref{tab:model_comparison exp 2}, demonstrate a clear distinction between the two tasks, but also between the models. 



\begin{figure}[ht]
    \centering
    \begin{subfigure}{0.23\textwidth}
        \includegraphics[width=\linewidth]{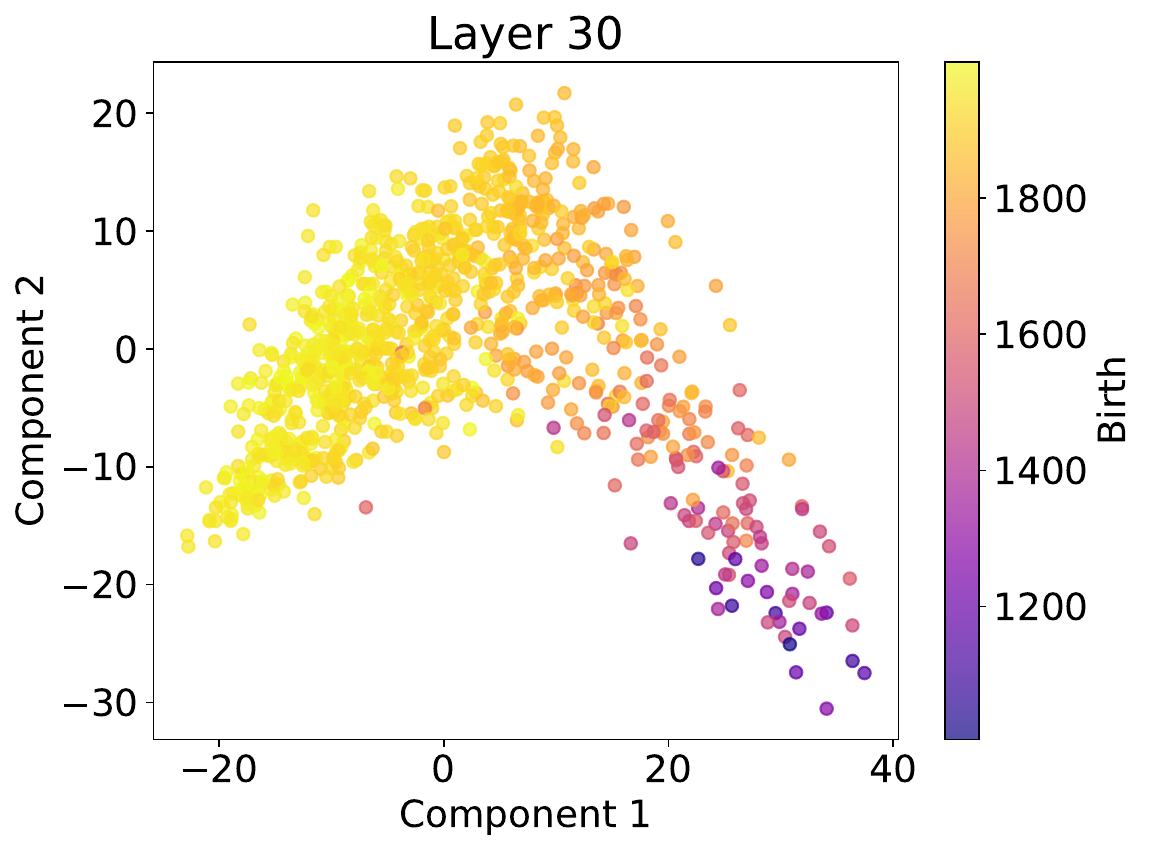}
    \end{subfigure}
    \hfill
    \begin{subfigure}{0.23\textwidth}
        \includegraphics[width=\linewidth]{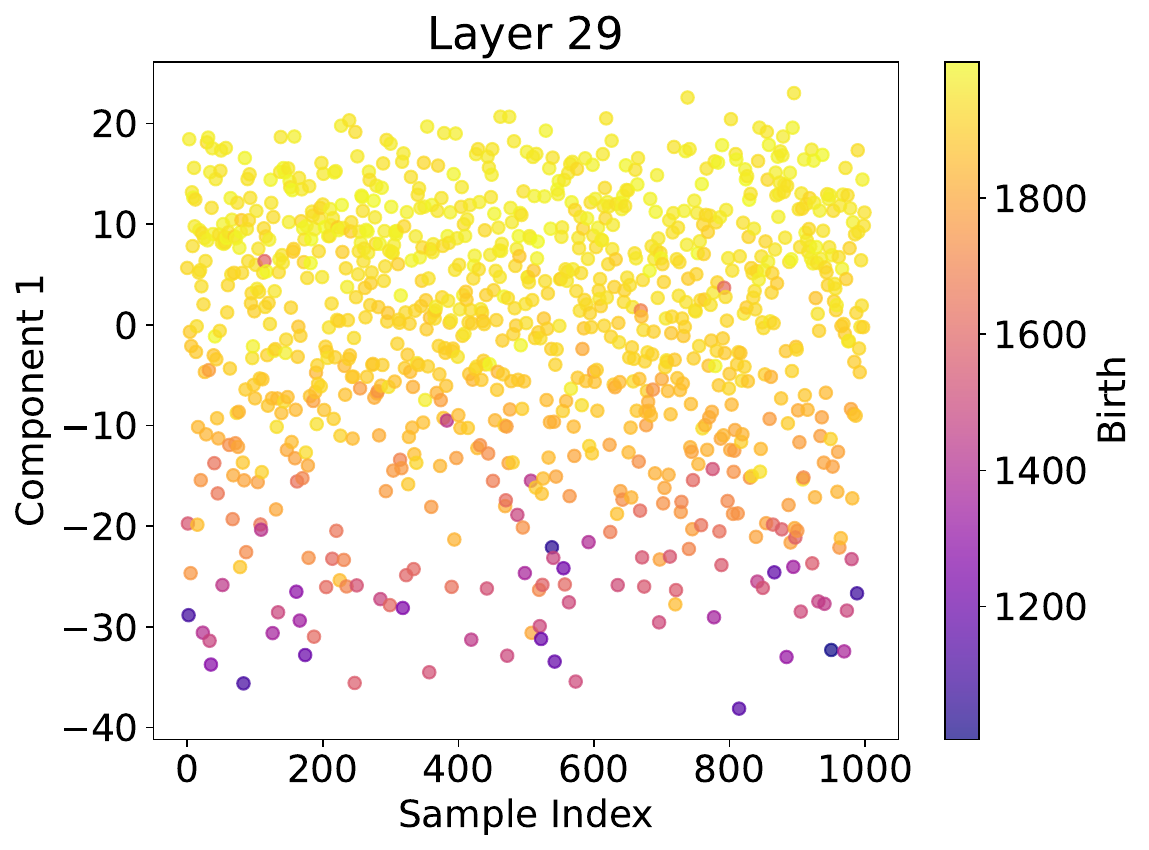}
    \end{subfigure}
    \hfill
    \begin{subfigure}{0.23\textwidth}
        \includegraphics[width=\linewidth]{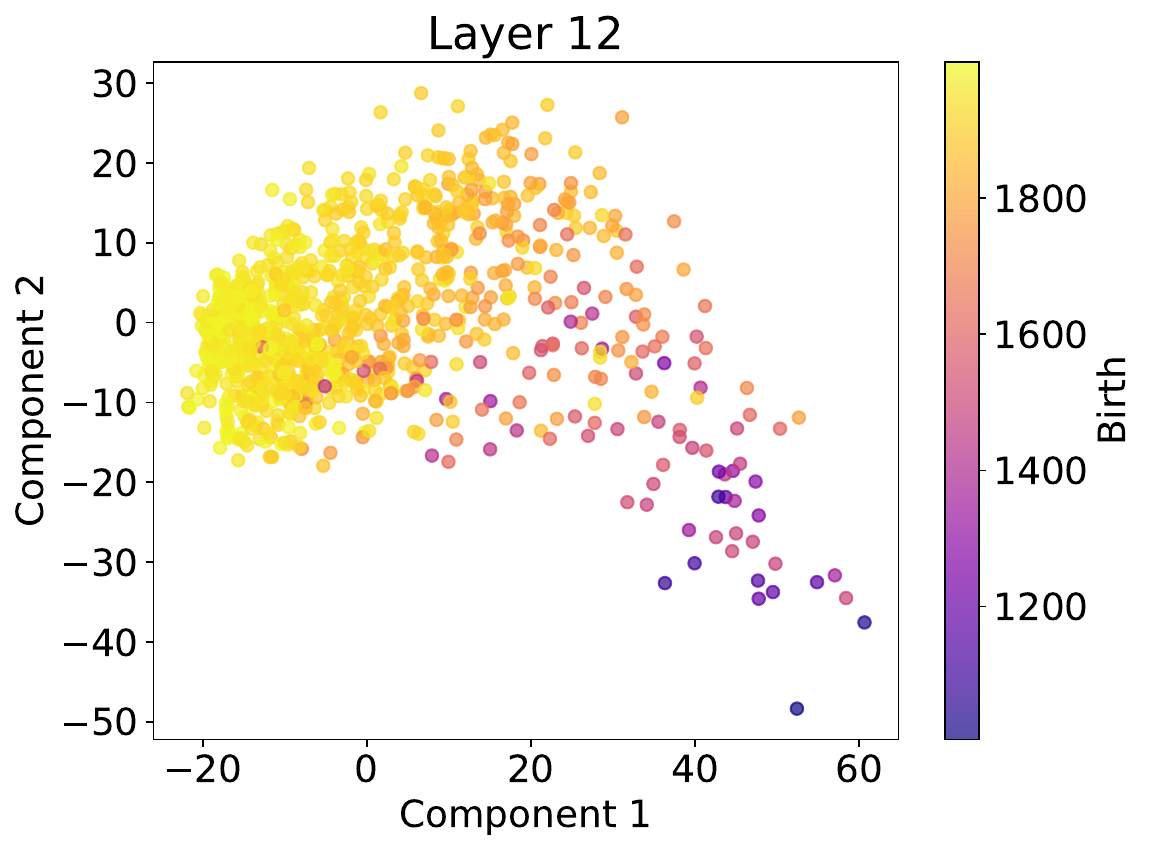}
    \end{subfigure}
    \hfill
    \begin{subfigure}{0.23\textwidth}
        \includegraphics[width=\linewidth]{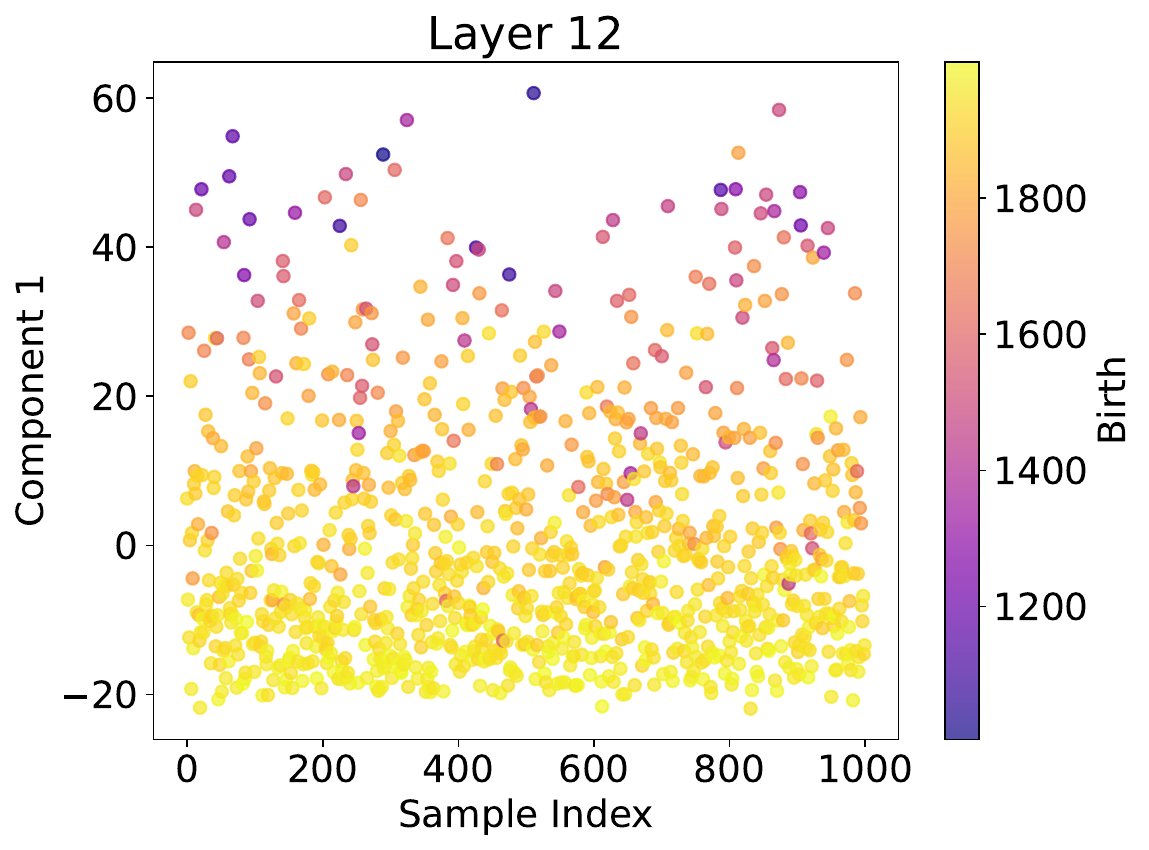}
    \end{subfigure}
    \caption{Visualization of PLS models trained on Llama-3.1-8B (top row) and Llama-3.2-1B (bottom row) model activations to predict entities' birth years using one and two dimensional PLS respectively. Each subfigure represents the layer with the highest $R^2$ score for one- and two-component PLS models.}
    \label{fig:pls_model_comparison}
\end{figure}

For the \textit{birth year} task, model \textit{Llama-3.1-8B} exhibit strong trends, with high monotonicity ($\rho$) and  ($R^2$), while having low SRI ($\beta)$, hence high compression. This indicates that the internal representations of birth years are well-structured and predictive, aligning with our expectations for numerical encoding in LLMs.
On the other side, \textit{Llama-3.2-1B}) shows low monotonicity score, hence the $\beta$ factor is not informative. We attribute the low monotonicity score to the non-structured internal representations of lower birth years, as can be seen in Figure \ref{fig:pls_model_comparison}.

For the \textit{population size} task, both models display weaker monotonicity and lower $R^2$, suggesting population sizes are encoded less systematically. Unlike birth years, population figures are more context-dependent, influenced by geopolitical changes, reporting inconsistencies, and approximate expressions in text. Consequently, the low monotonicity makes the scaling ratio $\beta$ unreliable.

Finally Figure \ref{fig:pls_model_comparison} shows the examples of one and two PLS projections for two models, for birth-year dataset.

\begin{table}[ht]
\begin{footnotesize}
    \centering
    \resizebox{.48\textwidth}{!}{ 
    \renewcommand{\arraystretch}{1.2}
    \setlength{\tabcolsep}{3pt}
    \begin{tabular}{lccccc}
        \toprule
        \textbf{Model} & \textbf{Dataset} & \textbf{Layer} & $\rho$ & $\beta$ & $R^2$ \\
        \midrule
        \multirow{2}{*}{Llamba 3.1 8B} & Birth & 29 & 0.84 & 0.50 & 0.63 \\
         & Population & 9 & 0.63 & 2.48 & 0.08 \\
        \midrule
        \multirow{2}{*}{Llamba 3.2 1B} & Birth & 12 & 0.03 & 0.72 & 0.61 \\
         & Population & 10 & 0.62 & 2.69 & 0.10 \\
        \bottomrule
    \end{tabular}
    }
\end{footnotesize}
\caption{Results for Llamba models evaluated on the Birth and Population datasets. Results are reported for the layer with the highest $R^2$, highlighting the relationship between scaling rate ($\beta$), monotonicity ($\rho$), and model performance.}
\label{tab:model_comparison exp 2}
\end{table}

%% file: sections/conclusion.tex
Inspired by the logarithmic compression in human numerical cognition, we investigate whether LLMs encode numerical values analogously. By analyzing hidden states across layers, we employ dimensionality reduction techniques (PCA and PLS) and geometric regression to test for two key properties: (1) order preservation and (2) sublinear compression, where distances between consecutive numbers decrease as values increase. Our results reveal that while both PCA and PLS identify numerical representations in a linear subspace, only PCA captures systematic sublinearity. This indicates that linear probes like PLS, which optimize for covariance with the target, may obscure the underlying non-uniform structure. Our findings suggest that LLMs encode numerical values with structured compression, akin to the human mental number line, but this is only detectable through methods like PCA that preserve geometric relationships.

%% file: sections/appendix.tex
\section{Experimental details}
 \label{app:exp}
All experiments were performed using an {NVIDIA A6000 GPU} for accelerated computation. The models were implemented in {Python} and imported from Huggingface with {PyTorch}, and standard libraries like {NumPy} and {Matplotlib} were used for data processing and visualization. We evaluated the following models:

\begin{table}[h!]
\centering
\begin{footnotesize}
\begin{tabular}{l|l|l}
\hline
\textbf{Model} & \textbf{Variants} &Ref. \\ \hline
Pythia & 2.8B &\cite{touvron2023llama1} \\ \hline
LLaMA & 2.7B, 3.1-8B, 3.2-1B & \cite{touvron2023llama1}\\ \hline
GPT-2 & Large-1.5B &\cite{radford2019language} \\ \hline
Mistral & 7B &\cite{jiang2024identifying}\\ \hline
\end{tabular}
\caption{Models evaluated in the experiments.}
\label{tab:models}
\end{footnotesize}
\end{table}

Whenever possible, results were reported as the average of three runs, along with the standard deviation (std). For experiments where repeated runs were not feasible, the random seed was fixed to {42} to ensure reproducibility. 
\section{Additional experiments}

\subsection{Layer-wise PLS analysis}
\begin{figure}[ht]
    \centering
    \begin{subfigure}[b]{0.23\textwidth}
        \centering
    \includegraphics[width=\textwidth]{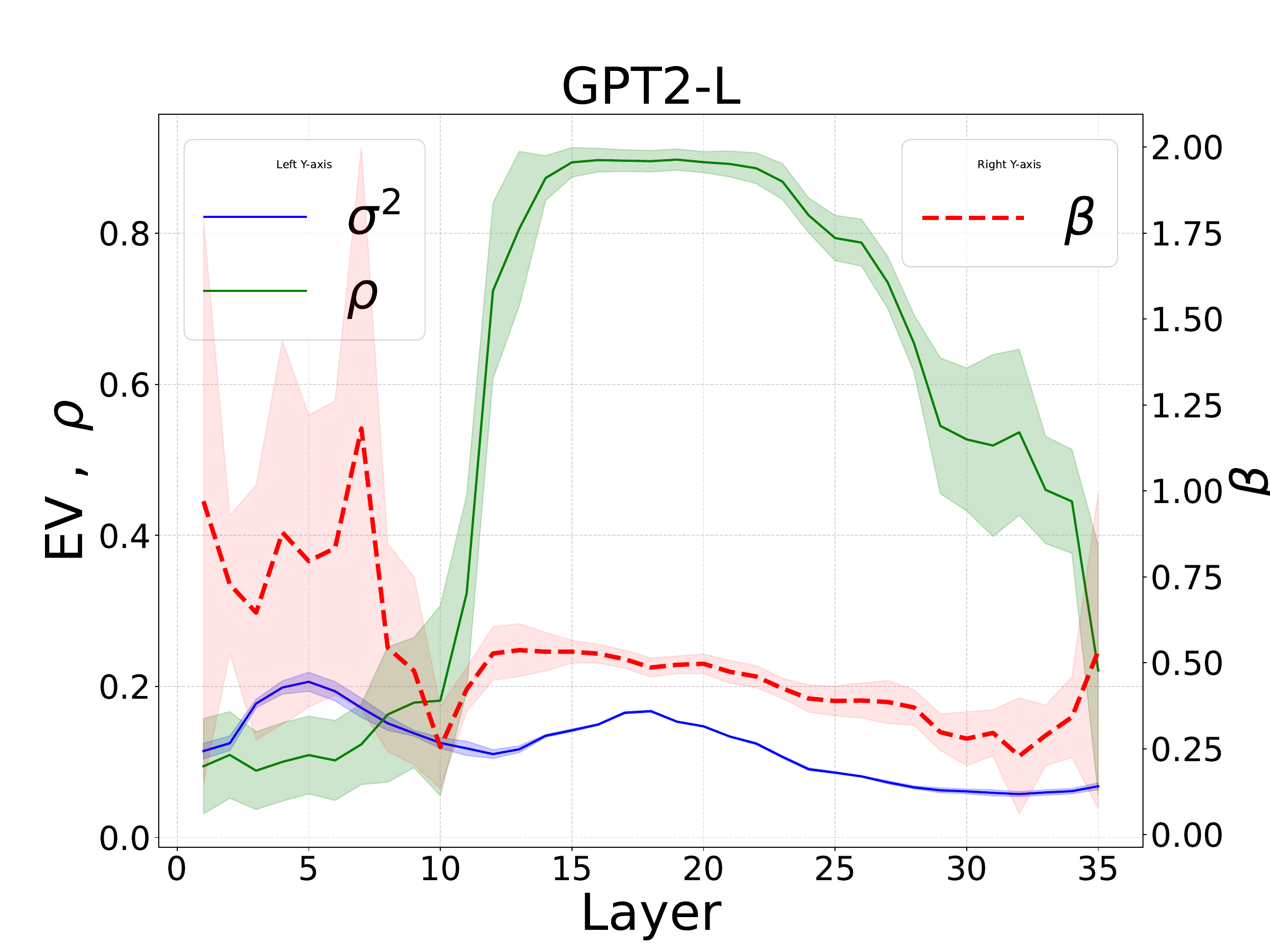}
    \end{subfigure}
    \hfill
    \begin{subfigure}[b]{0.23\textwidth}
        \centering
        \includegraphics[width=\textwidth]{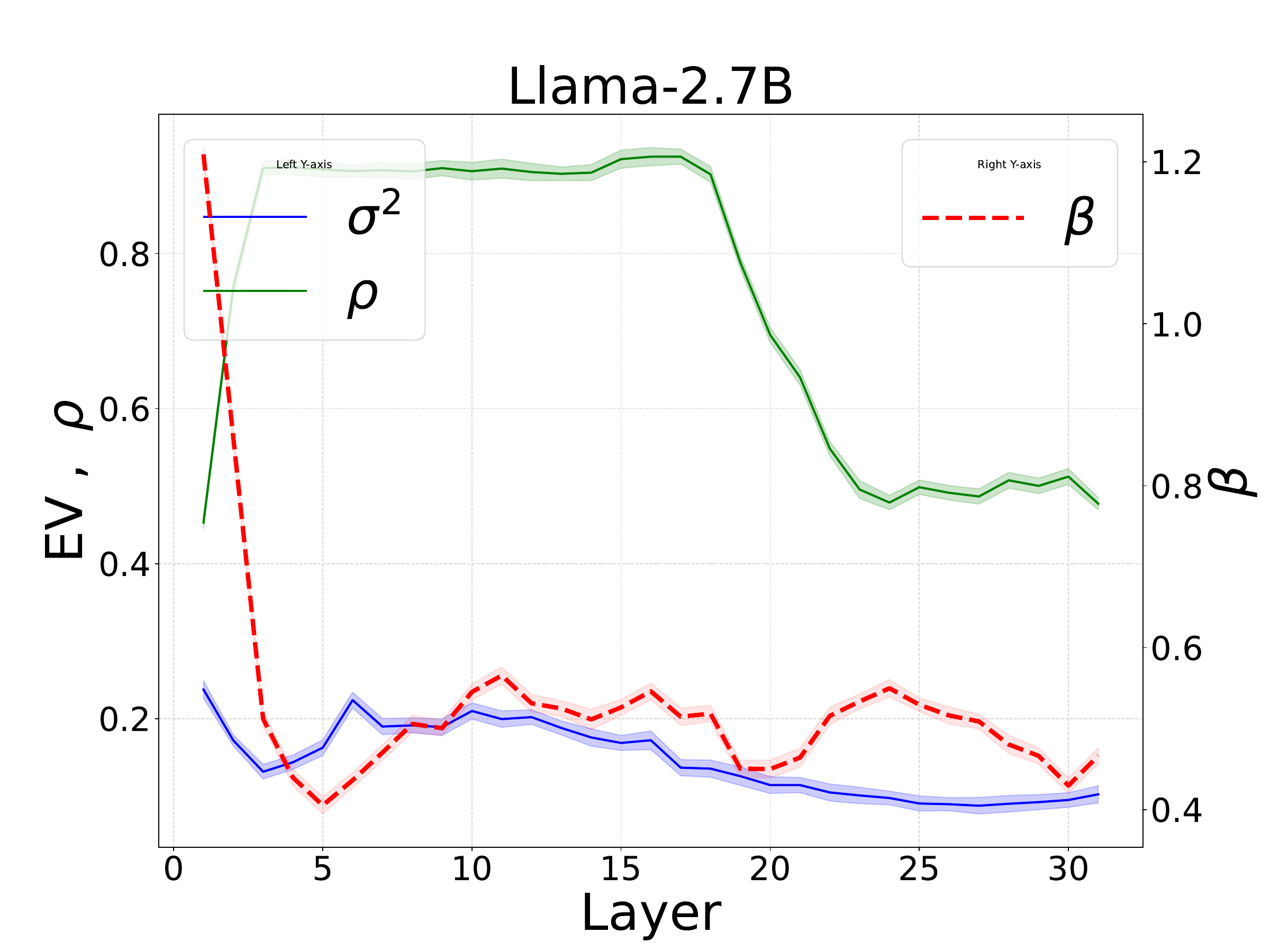}
    \end{subfigure}
    
    \vspace{0.3cm}  
    
    \begin{subfigure}[b]{0.23\textwidth}
        \centering
        \includegraphics[width=\textwidth]{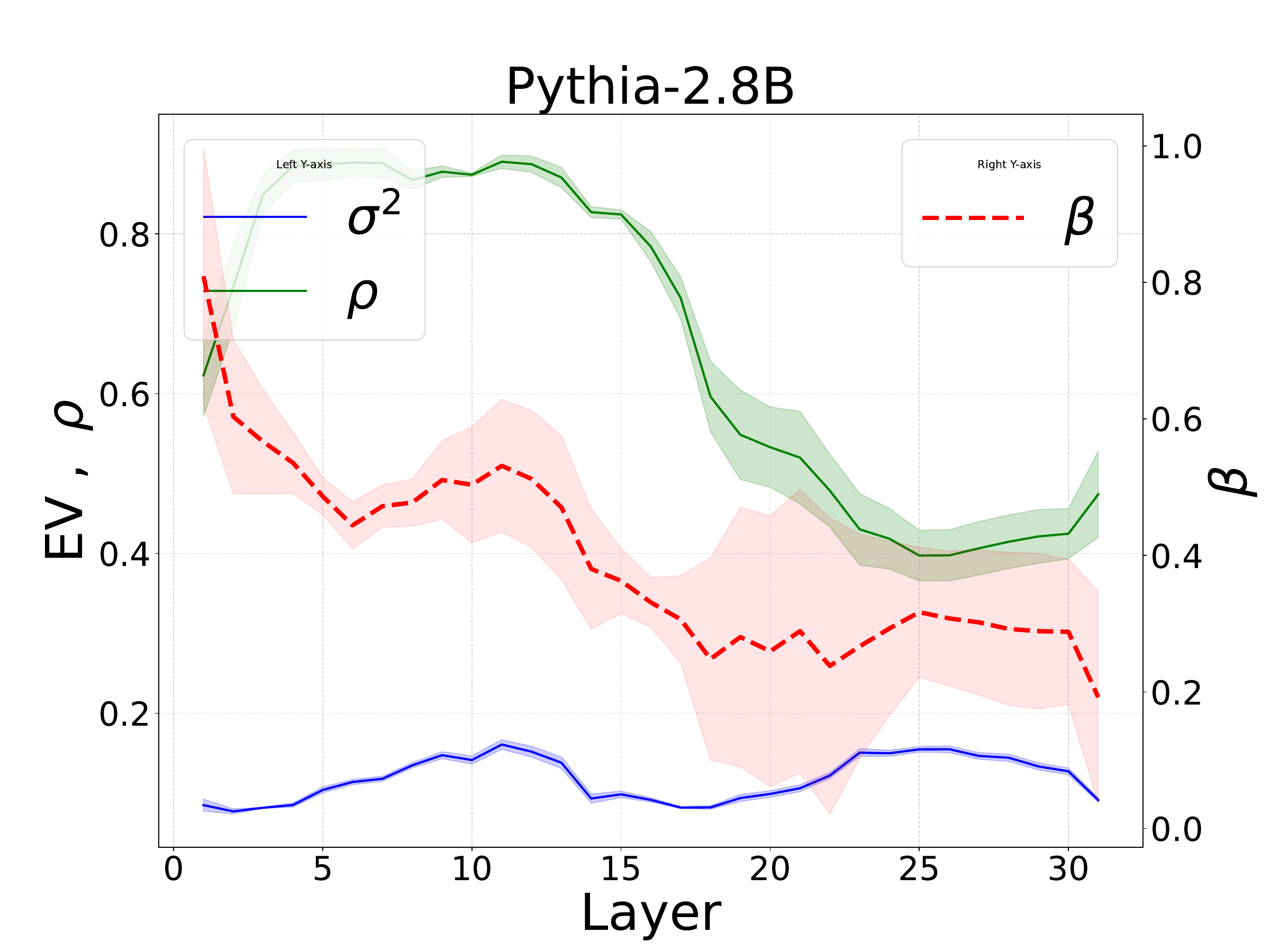}
    \end{subfigure}
    \hfill
    \begin{subfigure}[b]{0.23\textwidth}
        \centering
        \includegraphics[width=\textwidth]{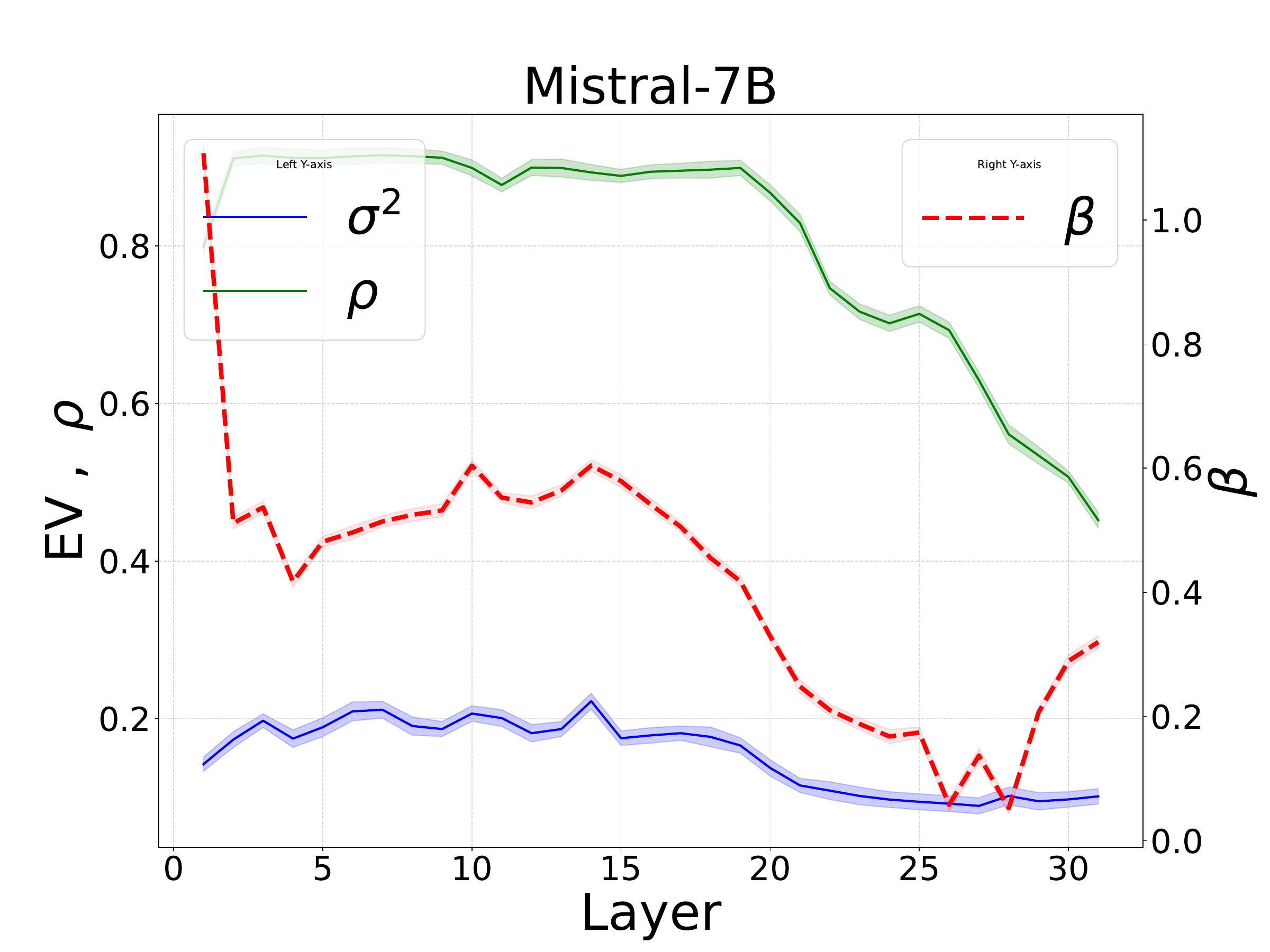}
    \end{subfigure}

\caption{Layer-wise analysis of four models on letters groups, showing explained variance ($\sigma^2$), monotonicity ($\rho$), and Scaling Rate Index  ($\beta$).}    
\label{fig:metrics}
\end{figure}
\subsection{Birth year  and population datasets projections in all layers}

\begin{figure}
    \centering
    \includegraphics[width=0.45\textwidth]{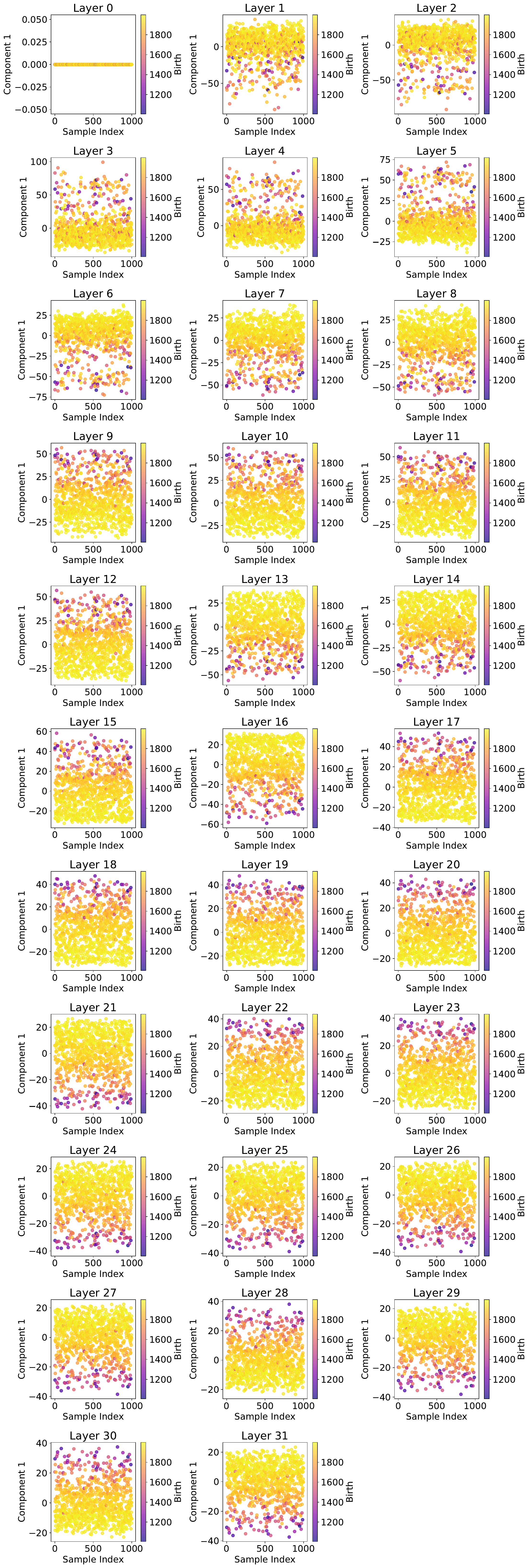}
    \caption{One component PLS model trained on Llama-3.1-8B instruct model activations to predict entities' birth year.}
    \label{fig:one-component pls-birth task}
\end{figure}

\begin{figure}
    \centering
    \includegraphics[width=0.45\textwidth]{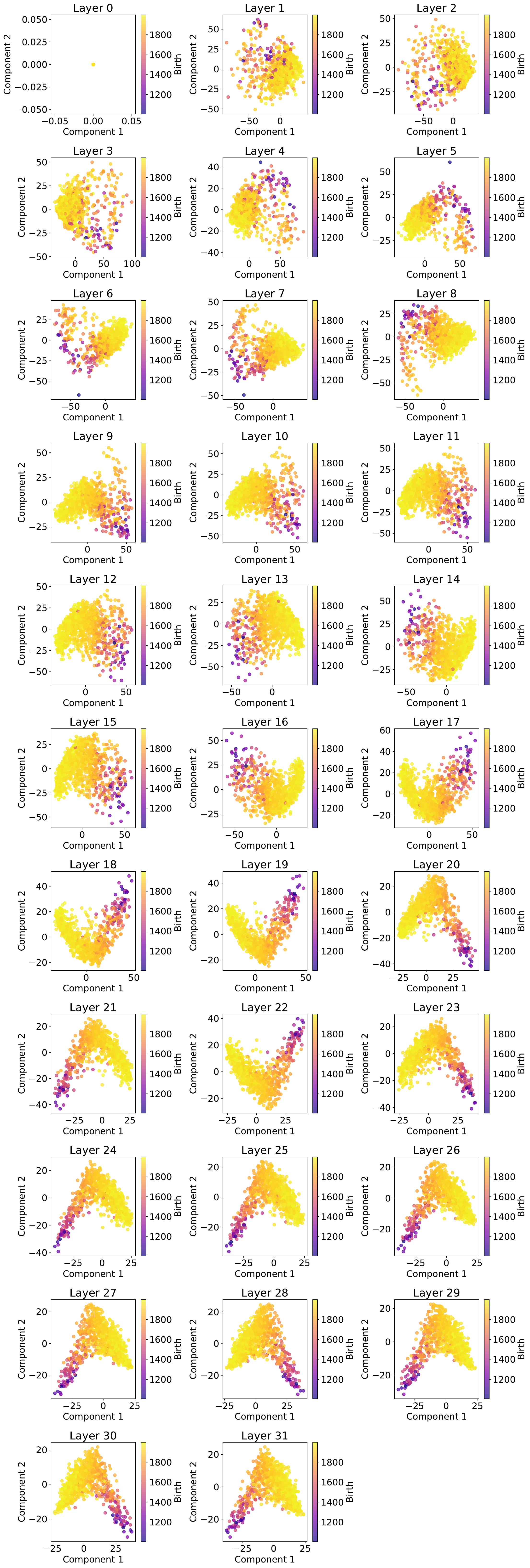} 
    \caption{Two components PLS model trained on Llama-3.1-8B instruct model activations to predict entities' birth year.}
    \label{fig:two-components pls-birth task}
\end{figure}

\begin{figure*}
    \centering
    \includegraphics[width=0.9\linewidth]{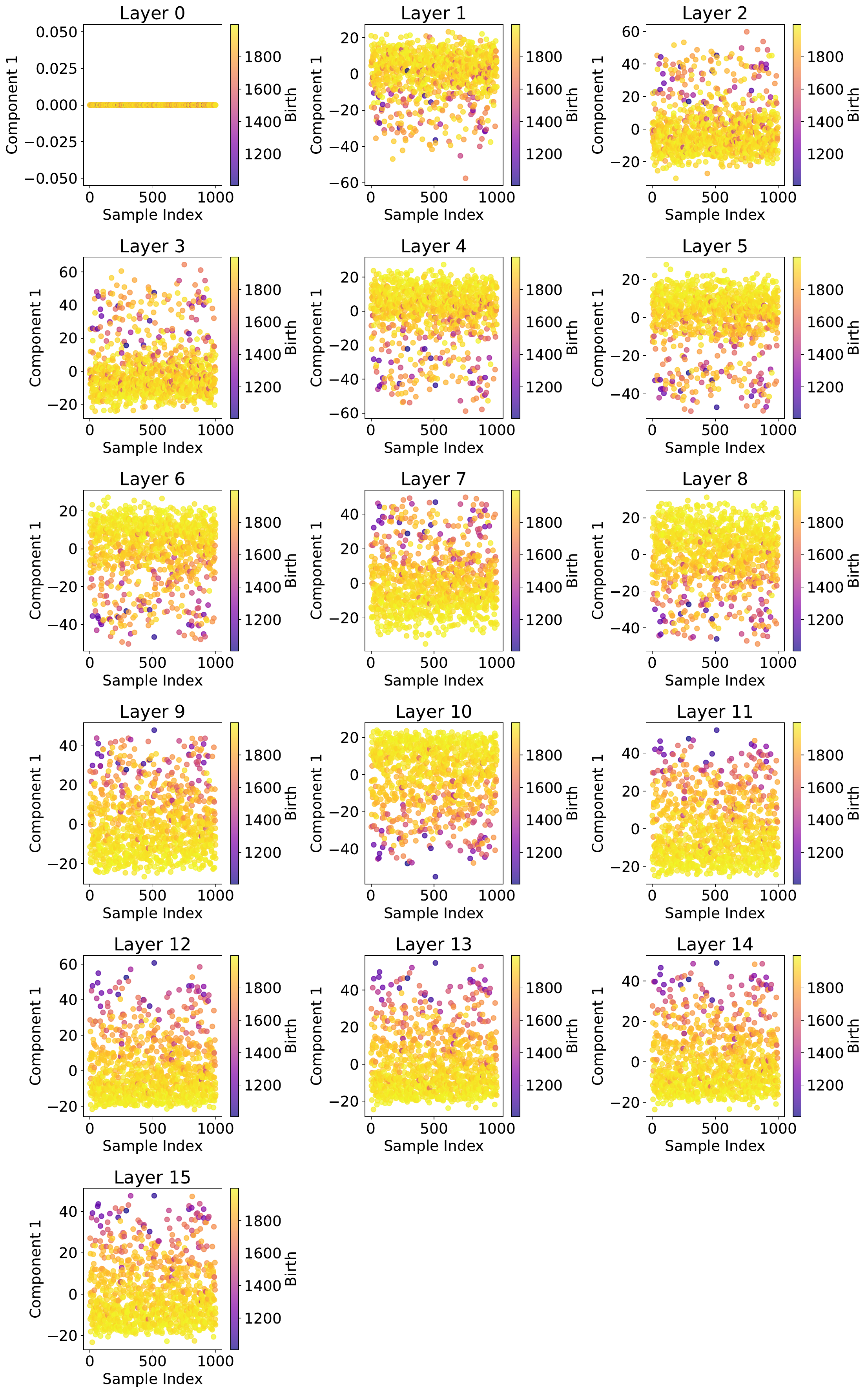}
    \caption{One component PLS model trained on Llama-3.2-1B instruct model activations to predict entities' birth year.}
    \label{fig:1B_one-component pls-birth task}
\end{figure*}

\begin{figure*}
    \centering
    \includegraphics[width=0.9\textwidth]{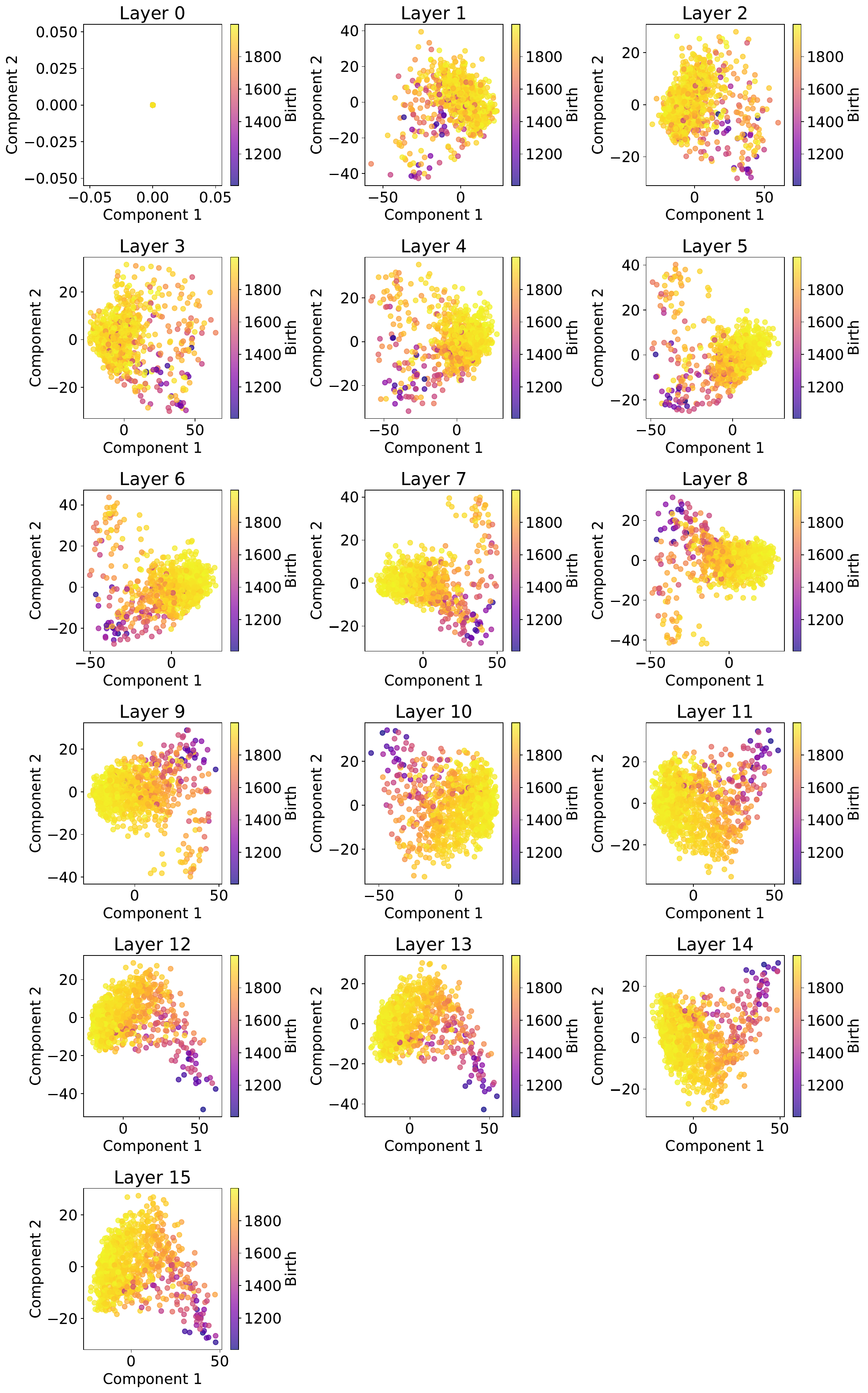}
    \caption{Two components PLS model trained on Llama-3.2-1B instruct model activations to predict entities' birth year.}
    \label{fig:1B-two-components pls-birth task}
\end{figure*}

\begin{figure}
    \centering
    \includegraphics[width=0.45\textwidth]{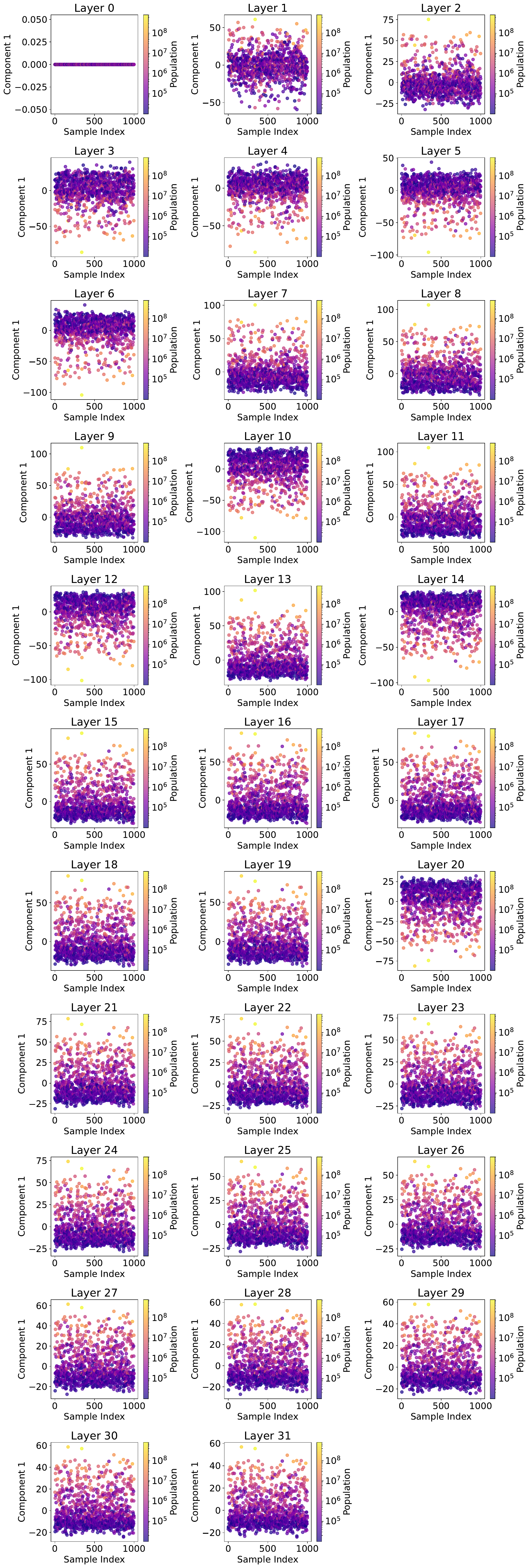}
    \caption{One component PLS model trained on Llama-3.1-8B instruct model activations to predict entities' population size.}
    \label{fig:one-component pls-population task}
\end{figure}

\begin{figure}
    \centering
    \includegraphics[width=0.45\textwidth]{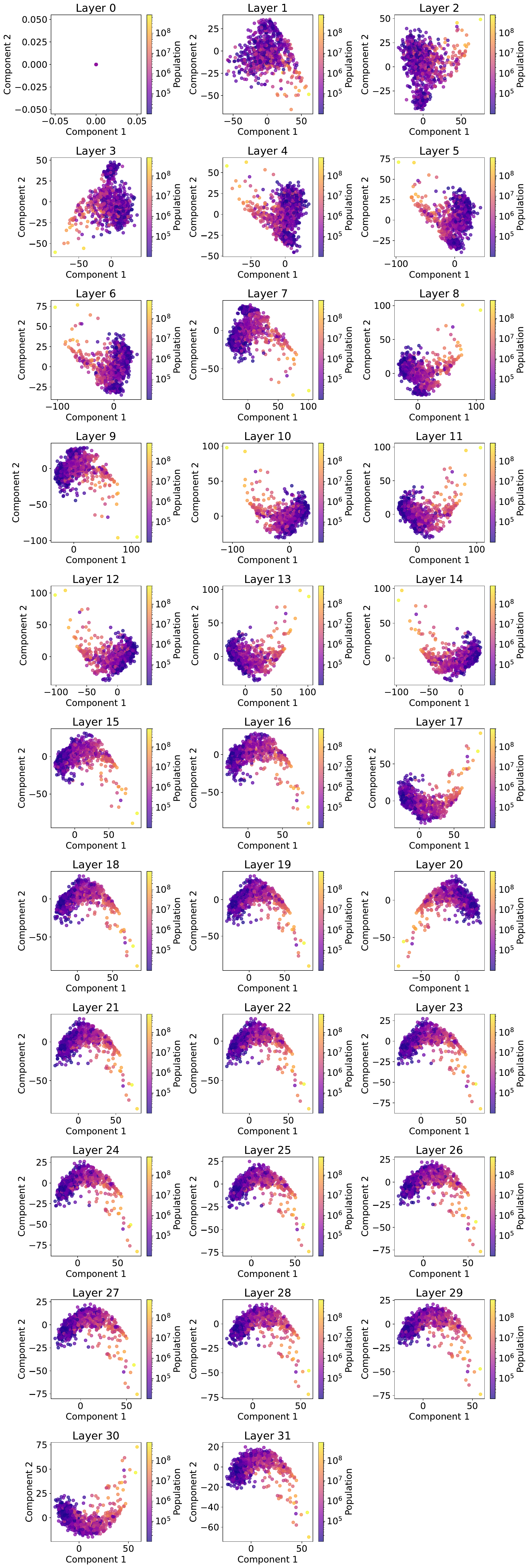}
    \caption{Two components PLS model trained on Llama-3.1-8B instruct model activations to predict entities' population size.}
    \label{fig:two-components pls-population task}
\end{figure}

\begin{figure*}
    \centering
    \includegraphics[width=0.9\textwidth]{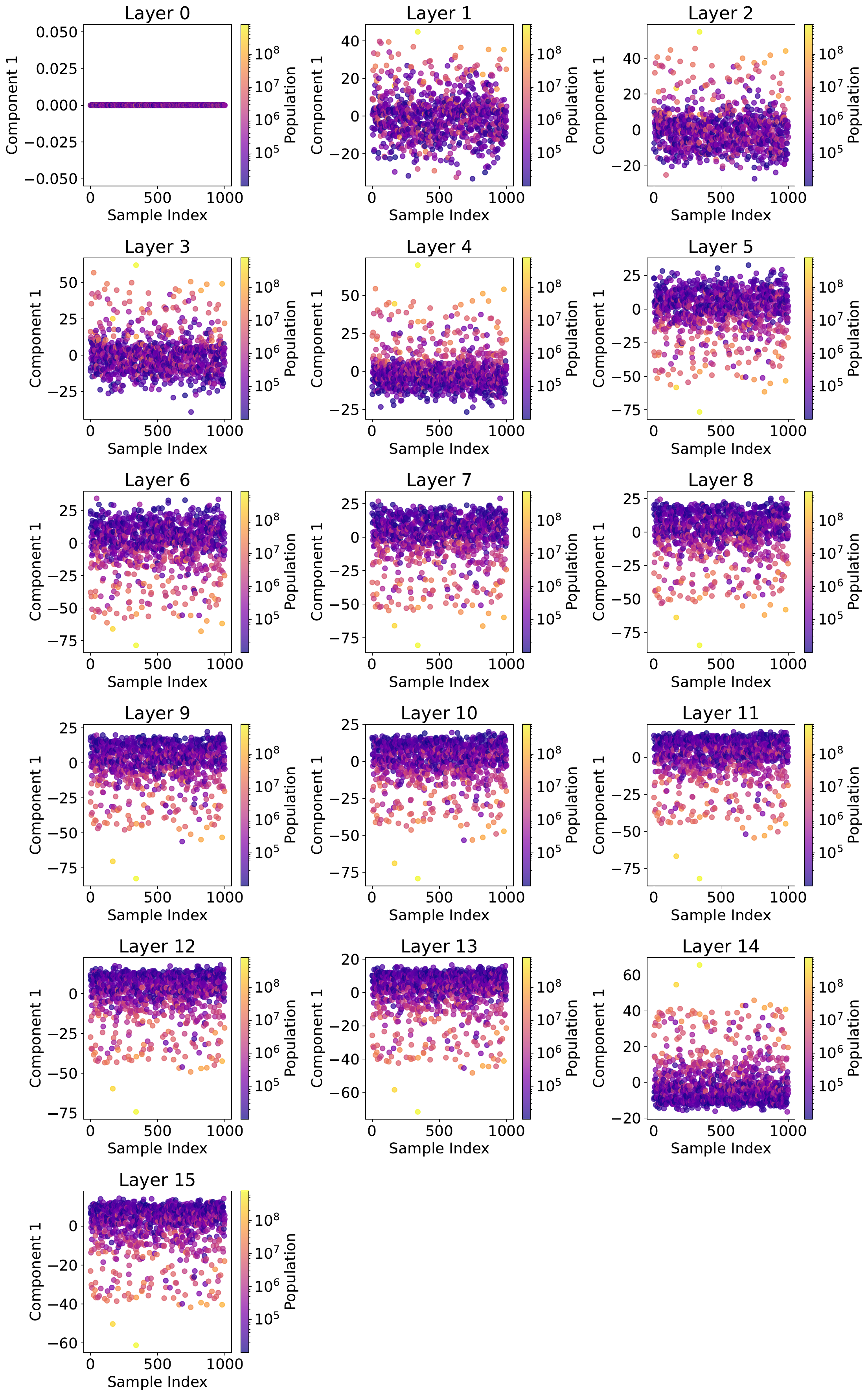}
    \caption{One component PLS model trained on Llama-3.2-1B instruct model activations to predict entities' population size.}
    \label{fig:1B-one-components pls-population task}
\end{figure*}
\begin{figure*}
    \centering
    \includegraphics[width=0.9\textwidth]{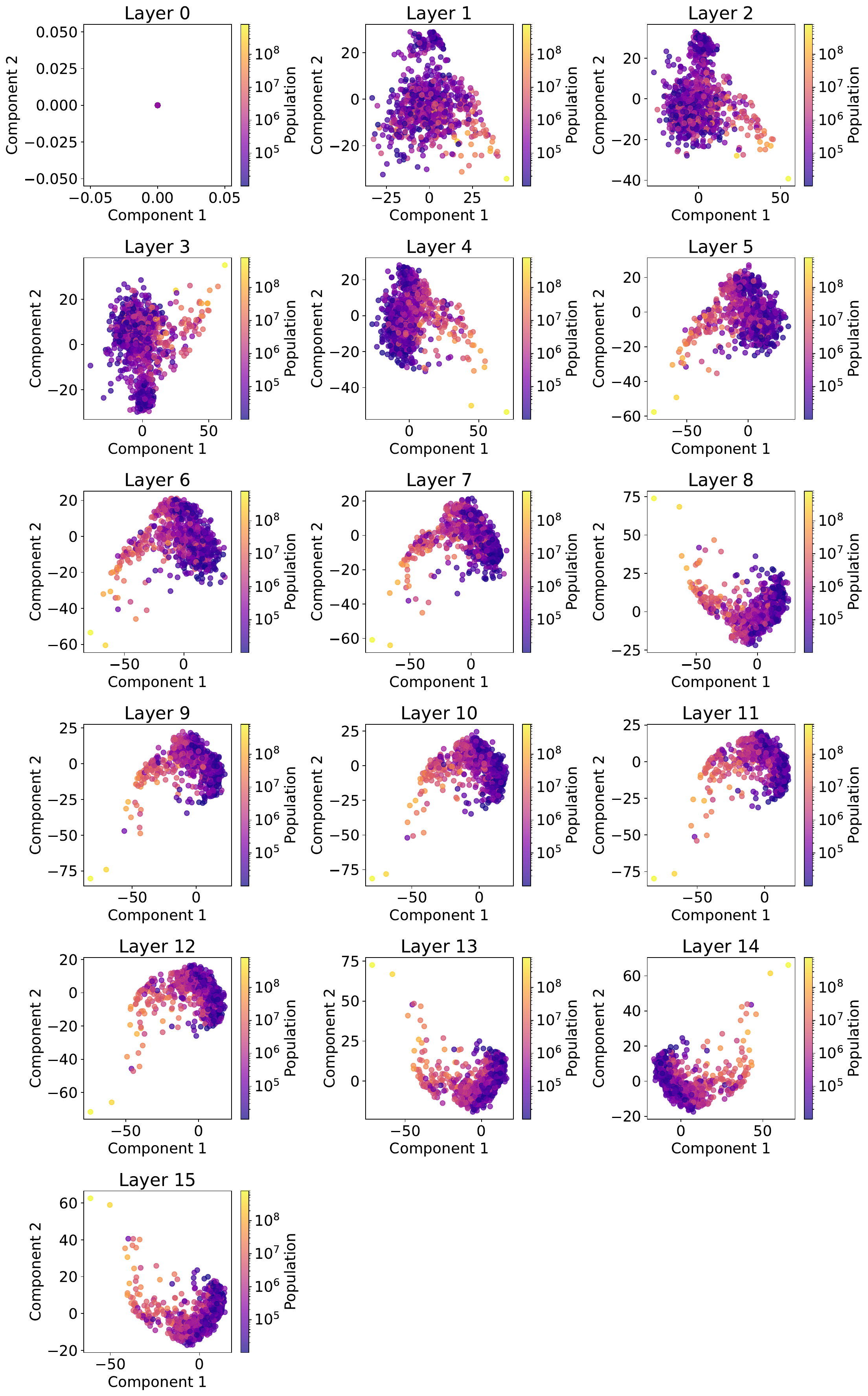}
    \caption{Two component PLS model trained on Llama-3.2-1B instruct model activations to predict entities' population size.}
    \label{fig:1B-two-components pls-population task}
\end{figure*}